\documentclass{article}

\usepackage{PRIMEarxiv}

\usepackage[T1]{fontenc}
\usepackage[utf8]{inputenc}
\usepackage{amsmath,amssymb}
\usepackage{mathptmx}
\usepackage{graphicx}
\usepackage{float}
\usepackage{placeins}
\usepackage{booktabs}
\usepackage{longtable}
\usepackage{microtype}
\usepackage{lineno}
\usepackage{xcolor}
\usepackage[small,labelfont=bf]{caption}
\usepackage[ruled,vlined,linesnumbered]{algorithm2e}
\usepackage{listings}
\usepackage[scaled=0.86]{zi4}
\usepackage{tcolorbox}
\tcbuselibrary{listings,breakable,skins}
\usepackage[numbers,sort&compress,square]{natbib}
\usepackage[hidelinks]{hyperref}
\hypersetup{
  pdftitle={Training-Free Agentic Computer Vision for Structural Component Detection in 2D Structural Framing Plans},
  pdfauthor={Mohammad Talebi-Kalaleh; Qipei Mei},
  pdfsubject={Training-free structural plan-to-model conversion},
  pdfkeywords={structural framing plans, finite-element model drafting, vector geometry, vision-language agents, benchmark}
}

\graphicspath{{figures/}}

\newcommand{\pdbench}{PD-50}
\newcommand{\pdtest}{PD-50-T}

\definecolor{boxnavy}{HTML}{1F4E79}
\definecolor{boxteal}{HTML}{2A8C8C}
\definecolor{boxframe}{HTML}{5A6872}
\definecolor{prompthead}{HTML}{D8E8F0}
\definecolor{schemahead}{HTML}{DCECE8}
\definecolor{codebg}{HTML}{F7F8FA}
\definecolor{codetext}{HTML}{24292E}
\definecolor{codenum}{HTML}{8B95A1}
\definecolor{lstkey}{HTML}{2E4374}
\definecolor{lstcom}{HTML}{55736E}
\definecolor{lststr}{HTML}{8A5A2C}
\lstdefinestyle{prompt}{
  basicstyle=\ttfamily\footnotesize\color{codetext},
  backgroundcolor=\color{codebg},
  numbers=left,
  numberstyle=\sffamily\scriptsize\color{codenum},
  numbersep=8pt,
  stepnumber=1,
  numberblanklines=false,
  xleftmargin=18pt,
  breaklines=true,
  breakindent=12pt,
  breakatwhitespace=true,
  columns=fullflexible,
  keepspaces=true,
  showstringspaces=false,
  lineskip=1pt,
  aboveskip=0pt,
  belowskip=0pt,
  moredelim=[is][\color{lstkey}\bfseries]{@@}{@@},
  moredelim=[is][\color{lstcom}]{§§}{§§},
}
\lstdefinestyle{json}{
  style=prompt,
  morestring=[b]",
  stringstyle=\color{lststr},
}
\lstdefinestyle{promptcompact}{
  style=prompt,
  basicstyle=\ttfamily\scriptsize\color{codetext},
  lineskip=0pt,
}

\newtcolorbox[auto counter]{promptbox}[2][]{
  colback=white,
  colframe=boxframe,
  colbacktitle=prompthead,
  coltitle=codetext,
  fonttitle=\sffamily\bfseries\small,
  title={Prompt~\thetcbcounter\quad #2},
  enhanced,
  boxrule=0.8pt,
  arc=3mm,
  outer arc=3mm,
  lefttitle=3.5mm, righttitle=3.5mm,
  toptitle=1.5mm, bottomtitle=1.5mm,
  breakable,
  left=3.5mm, right=3.5mm, top=2.8mm, bottom=2.4mm,
  before skip=10pt, after skip=8pt,
  #1
}
\newtcolorbox[use counter from=promptbox]{schemabox}[2][]{
  colback=white,
  colframe=boxframe,
  colbacktitle=schemahead,
  coltitle=codetext,
  fonttitle=\sffamily\bfseries\small,
  title={Schema~\thetcbcounter\quad #2},
  enhanced,
  boxrule=0.8pt,
  arc=3mm,
  outer arc=3mm,
  lefttitle=3.5mm, righttitle=3.5mm,
  toptitle=1.5mm, bottomtitle=1.5mm,
  breakable,
  left=3.5mm, right=3.5mm, top=2.8mm, bottom=2.4mm,
  before skip=10pt, after skip=8pt,
  #1
}

\SetKwInput{KwCfg}{Config}
\SetKwComment{tcp}{$\triangleright$\ }{}
\SetAlCapSkip{4pt}

\title{Training-Free Agentic Computer Vision for Structural Component
Detection in 2D Structural Framing Plans}

\author{
  Mohammad Talebi-Kalaleh \\
  Department of Civil and Environmental Engineering \\
  University of Alberta \\
  \texttt{talebika@ualberta.ca} \\
  \And
  Qipei Mei \\
  Department of Civil and Environmental Engineering \\
  University of Alberta \\
  \texttt{qipei.mei@ualberta.ca} \\
}

\date{}

\begin{document}

% \linenumbers   % journal submission only; arXiv preprint has no line numbers
\maketitle

% Kept within 1,920 characters so the manuscript abstract and the submission
% portal field are identical text.
\begin{abstract}
Converting structural framing plans into editable finite-element model drafts
is labor-intensive and susceptible to transcription errors. Existing
building-component recognition systems generally depend on task-specific
neural detectors, whereas language-model agents in structural engineering
typically operate on text or model data rather than on drawings. To the authors'
knowledge, this work is the first to apply an agentic vision-language layer to
structural-component detection and model drafting from framing-plan PDFs without
task-specific detector training or fine-tuning. A deterministic stage extracts
geometric primitives, estimates scale by dimension-ratio consensus, recognizes
five entity classes using an explicit drafting grammar, and assembles an
editable layout. The agentic stage constrains typed corrections through
deterministic candidates, operation-specific admission tests, change-level
review, and fail-closed transactions. Evaluation used an author-generated
benchmark of 100 plans, divided equally between a development half used for all
rule revisions and a seed-disjoint held-out half generated after the rules were
frozen and evaluated once. All scores are end-to-end results for the complete
framework on the held-out half. Scale estimates were within 0.1\% of the
generator reference for every drawing. Recall and precision were 0.922/0.997
for columns, 0.886/0.990 for beams, 1.000/1.000 for walls, 1.000/1.000 for
braces, and 1.000/0.964 for openings. A controlled study repeated two
corruptions three times on three development drawings. Calibration passed all
nine trials, whereas member repair satisfied every strict end-state criterion
in five of nine trials. Because both benchmark halves share a generator, the
evaluation does not address independently drafted plans, raster input,
analytical connectivity, or solver validation.
\end{abstract}


\section{Introduction}
\label{sec:intro}

Structural framing plans encode the column grid, member layout, section
designations, wall and brace locations, and slab boundaries for individual
building floors
\citep{zhao2020deep,zhao2021reconstructing}. When an engineer needs a
finite-element model of that building, whether for a retrofit study, peer
review, or progressive-collapse check, the plan must be translated into nodes,
elements, sections, and supports. In current practice, this translation is
largely manual: coordinates are read from dimension strings, members are
redrawn in analysis software, and section labels are entered again.
\citet{gimenez2015review} characterized manual three-dimensional model creation
from drawings as complex and time-consuming and found no complete automatic
reconstruction pipeline. Even when a building information model is available,
deriving a geometrically faithful analysis model requires dedicated tools
\citep{hasan2019bimfem}.

Two research communities have approached the drawing-understanding problem
from opposite directions. Document-analysis research has moved from
rule-based interpretation of architectural drawings
\citep{dosch2000complete,mace2010system,ahmed2011improved} to deep networks
that parse raster floor plans \citep{liu2017raster,zeng2019deep,
kalervo2019cubicasa} and, more recently, to graph and transformer models
that operate directly on CAD vector primitives
\citep{fan2021floorplancad,zheng2022gatcadnet,fan2022cadtransformer,
yang2023vectorfloorseg,liu2024symbol}. These methods target architectural
semantics, such as rooms, doors, and furniture symbols, and report scores in
image or primitive space. Structural-model reconstruction imposes additional
requirements: member endpoints must be reconciled into a connected topology,
and an incorrect scale propagates to every coordinate. In parallel, the
language-model community has produced agents that reason, call tools, and
review their own output \citep{yao2023react,schick2023toolformer,
madaan2023selfrefine,zheng2023judge}, and these agents have begun to automate
structural engineering workflows from textual descriptions
\citep{liang2025structural,liang2025masse,du2026text2bim}. A recent review of
134 studies identifies model generation and design checking as major areas of
artificial-intelligence (AI)-enabled structural design automation
\citep{xie2025aiapplications}. Despite this progress, vision-language models
remain unreliable for low-level visual reasoning tasks relevant to plan review,
including counting and spatial relations. Across seven synthetic tasks, four
vision-language models averaged 58.07\% accuracy
\citep{rahmanzadehgervi2024blind}, and
ungrounded generation is subject to hallucination
\citep{ji2023hallucination}.

The proposed method assigns geometric measurement to deterministic procedures
before invoking vision-language interpretation. The deterministic layer handles
quantities defined by explicit geometric predicates, including primitive
extraction, scale consensus, symbol signatures, bearing topology, and region
analysis. The refinement layer addresses semantic questions, such as whether
unexplained linework represents a missed girder or a title-block underline,
through typed operations.
Constructive edits must satisfy paper-space ink or glyph tests, whereas
deletions, moves, attribute changes, and calibration retain
operation-specific guards and vision-model review rather than a general
geometric proof. This allocation restricts the decisions delegated to the
vision-language model without implying that every accepted semantic judgment is
correct. The framework uses neither a task-specific learned detector nor
task-specific fine-tuning; extending the supported notation requires an
explicit rule and a new validation cycle
rather than model retraining. Section~\ref{sec:conclusions} therefore treats
published learned systems as related methods, not as commensurate baselines.

The work makes three contributions. First, it defines a deterministic
conversion pipeline from PDF primitives to an editable floor layout and a
three-dimensional finite-element model draft. The pipeline includes explicit
scale resolution, a structural drafting grammar, and topology-based model
assembly. Second, it introduces, to the authors' knowledge, the first agentic
vision-language architecture for building-component detection and structural
model drafting from drawings. Prior drawing-understanding systems train
task-specific neural detectors
\citep{zhao2020deep,fan2022cadtransformer,xie2025semisupervised}, and prior
language-model agents in structural engineering act on textual briefs,
scripts, or model data rather than the drawing image
\citep{liang2025structural,liang2025masse,du2026text2bim}. Within the proposed
architecture, the vision-language model can propose only typed operations
bounded by deterministic candidate generators, operation-specific admission
tests, change-level judging, and fail-closed handling of entity edits. The two
layers are designed and evaluated as a single system: deterministic geometry
supplies measured coordinates, while the agentic layer interprets designations,
materials, and calibration evidence that geometry alone cannot resolve. Third,
the work introduces a 100-drawing benchmark with exact generator ground truth.
The benchmark is divided equally into the \pdbench{} development half, which
informed rule and threshold revisions, and the seed-disjoint \pdtest{} held-out
half, which was generated after the detection rules were frozen and evaluated
once. Every reported detection recall and precision value is an end-to-end
result from the held-out half. The evaluation also reports disaggregated results
and repeated controlled corruption trials for the guarded refinement layer.

The remainder of the paper first situates the method within drawing analysis,
drawing-to-model reconstruction, and agentic structural engineering in
Section~\ref{sec:related}. Section~\ref{sec:whyhybrid} then explains the
division of responsibilities and presents the overall framework.
Sections~\ref{sec:deterministic} and~\ref{sec:agentic} specify the deterministic
and agentic layers, respectively. Section~\ref{sec:benchmark} describes the
benchmark and evaluation protocol, and Section~\ref{sec:results} reports the
results. Section~\ref{sec:conclusions} summarizes the findings, their practical
meaning, the study limitations, and the required next experiments.

\section{Related work}
\label{sec:related}

\subsection{Floor plan and CAD drawing analysis}

Early systems interpreted architectural drawings using manually specified
rules applied to vectorized primitives. \citet{dosch2000complete} combined vectorization,
symbol recognition, and cross-floor matching into a complete
scanned-drawing-to-3D pipeline. \citet{mace2010system} detected walls and rooms
through Hough-based line detection and recursive region decomposition, whereas
\citet{ahmed2011improved} separated walls from annotation by line thickness.
This classical line of work, surveyed by \citet{tombre1998analysis}, established
vectorization methods that remain in use \citep{hilaire2006robust}. Subsequent
statistical methods sought recognition across notation systems and drawing styles
\citep{delasheras2014statistical}.

Deep learning subsequently reframed the problem as pixel prediction. Raster-to-Vector
\citep{liu2017raster} recovered vector floor plans from images through
junction detection and integer programming; multi-task networks
\citep{zeng2019deep,kalervo2019cubicasa} segmented walls, openings, and
rooms on datasets such as CubiCasa5K. A later shift returned to the vector
domain. FloorPlanCAD \citep{fan2021floorplancad} released more than ten
thousand CAD drawings with primitive-level annotation and defined the
panoptic symbol-spotting task, addressed by graph attention over primitives
\citep{zheng2022gatcadnet}, primitive-token transformers
\citep{fan2022cadtransformer}, two-stream graph networks on roughcast plans
\citep{yang2023vectorfloorseg}, and point-based primitive representations
\citep{liu2024symbol}. Surveys document the same trend across engineering
diagrams generally \citep{morenogarcia2019new,jamieson2025towards,
khade2026comprehensive}. These studies report pixel accuracy, mean class
accuracy, intersection over union, room matching, or panoptic quality under
dataset-specific protocols. Section~\ref{sec:conclusions} explains why those
values cannot serve as a numerical baseline for the object-level metric used
here.

Recent work has also focused on architectural layouts that support structural
design automation. \citet{xie2025semisupervised} combined a hierarchical vision
transformer with consistency regularization to segment walls from limited
labeled data. \citet{xie2025retrieval} used transformer-based wall segmentation
to retrieve similar wood-frame layouts and identify drawing differences.
These studies demonstrate the value of learned wall representations for design
reuse, but their outputs remain pixel-level architectural wall masks rather than
scaled structural entities and connected analysis-model topology. The present
study addresses the downstream conversion from drawing entities to
structural-model topology while retaining wall detection as
one component of a broader structural schema.

The proposed method differs from this literature in three respects. Its target
entities are structural rather than architectural, and it constructs bearing
relations after symbol detection. It expresses coordinates in model units after
explicit scale resolution instead of retaining pixel coordinates, subject to
the evaluation tolerances in Section~\ref{sec:evaluation}. Finally, its
detection path uses explicit geometric rules rather than a fitted detector, so
individual rules can be tested during error analysis. This structure improves
traceability but does not establish structural correctness.

\subsection{Drawing-to-model and building information modeling reconstruction}

The reconstruction of building models from legacy drawings is commonly studied
under scan-to-building information modeling (scan-to-BIM). A review by
\citet{gimenez2015review} found no fully automatic system, and a subsequent
study demonstrated a semi-automatic pipeline for scanned plans
\citep{gimenez2016automatic}. \citet{lu2007automatic}
integrated recognized objects across multiple drawings in a set. Of greatest
relevance here, \citet{zhao2020deep} detected structural components in scanned
structural drawings with an object detector, and \citet{zhao2021reconstructing}
assembled grids, columns, and beams into an Industry Foundation Classes (IFC)
model through a hybrid image-processing and optical character recognition
(OCR) pipeline. These related systems process raster drawings and reconstruct
BIM geometry, and the methods of \citet{zhao2020deep,zhao2021reconstructing}
rely on task-specific object detection. By contrast, the proposed pipeline
begins with vector primitives when available, uses no task-specific learned
detector, resolves sections and materials from printed designations, and emits
an editable finite-element model draft with supports, releases, and slab
meshes. End-to-end structural-analysis validity is outside the present
evaluation.

\subsection{Language-model agents in structural engineering}

Large language models (LLMs) have entered construction and structural workflows
as code generators and multi-agent planners. The systematic review by
\citet{xie2025aiapplications} identifies language models and automated design
checking among the emerging directions for AI in structural design. LLM agents
translate textual
building descriptions into executable structural-analysis scripts
\citep{liang2025structural}, coordinate multi-agent workflows that automate
routine structural engineering tasks \citep{liang2025masse}, generate BIM
models from natural-language briefs \citep{du2026text2bim}, and check code
compliance against building information models \citep{madireddy2025compliance}.
Assessments of general-purpose models in construction report both the
opportunity and the reliability limits \citep{saka2024gpt}. On the
perception side, document transformers parse structured images end to end
\citep{kim2022donut,lee2023pix2struct}, and question-answering corpora over
piping and instrumentation diagrams show spatial and counting questions to
be the hard categories \citep{gupta2025pidqa}.

The agentic architecture draws on four ideas from the literature: interleaved
reasoning and acting \citep{yao2023react}, tool
delegation for operations the model cannot perform reliably
\citep{schick2023toolformer}, iterative self-refinement
\citep{madaan2023selfrefine}, and model-based judging of model output
\citep{zheng2023judge}, with verification applied at the level of individual
steps rather than final outcomes, in the spirit of process supervision
\citep{lightman2024verify}. Combining a symbolic scaffold with
learned components follows the neurosymbolic program articulated by
\citet{garcez2023neurosymbolic}. In the present architecture, this guard
discipline prevents the language model from writing the layout directly.
Instead, typed proposals pass operation-specific checks, and a separate judge
call can reject an individual change without discarding the remaining changes.
Deterministic seeds share this operation channel, but their use does not
constitute geometric proof of every accepted edit.

Across these research areas, the reviewed drawing-understanding systems for
building components rely on trained neural detectors, from convolutional parsers
\citep{liu2017raster,zhao2020deep} to graph and transformer models over vector
primitives
\citep{zheng2022gatcadnet,fan2022cadtransformer,yang2023vectorfloorseg,xie2025semisupervised,xie2025retrieval},
whereas the reviewed agentic language-model systems in structural engineering
consume textual briefs, scripts, or building models rather than the drawing
image itself
\citep{liang2025structural,liang2025masse,du2026text2bim,madireddy2025compliance}.
To the authors' knowledge, the literature contains no prior system that applies
an agentic vision-language layer to building-component detection and structural
model drafting directly from drawings. The proposed framework addresses this
gap.

\section{Proposed plan-to-model method}
\label{sec:whyhybrid}

\subsection{Division of responsibilities}

Three architectures are relevant to framing-plan interpretation: a
task-specific network trained on annotated drawings, a vision-language model
prompted with the drawing image, and a deterministic geometric pipeline. A
task-specific network can learn broad appearance variation, but it requires a
representative annotated corpus. A vision-language model can interpret symbols
and notes without task-specific fine-tuning, whereas deterministic geometry can
measure coordinates and apply traceable predicates. No single architecture
satisfies the combined requirements for metric accuracy, semantic
interpretation, and bounded state changes. The proposed framework therefore
assigns each task to the component that can constrain it most directly.

A vision-language model is useful for interpreting hatched bands, brace signs,
and designations such as W14$\times$90, but four limitations prevent it from acting
as the sole geometric extractor:

\begin{enumerate}
\item \textit{Coordinate fidelity.} A coordinate proposed from an image is an
estimate rather than a measurement from vector geometry. At a drawing scale of
1:100, a 0.3~m model-space displacement occupies only 3~mm on the printed page,
which is difficult to estimate reliably from a downsampled prompt image.
\item \textit{Geometric reasoning.} Intersection, collinearity, and parallel-line
counts resemble low-level visual tasks on which four tested models averaged
58.07\% across a seven-task synthetic benchmark
\citep{rahmanzadehgervi2024blind}. Framing-plan interpretation invokes the same
predicates when identifying joists, beam endpoints, and crossing diagonals.
\item \textit{Dense enumeration.} Individual drawings contain as many as
85 beams. Because complete enumeration combines spatial search and counting,
the method does not assign it to a single model call; \citet{gupta2025pidqa}
similarly distinguishes simple counting from spatial-counting questions.
\item \textit{Independent verification.} A fluent response does not establish
that a proposed member exists \citep{ji2023hallucination}. A fabricated
entity can appear complete in the resulting model, whereas a missed entity
remains visible in the review overlay.
\end{enumerate}

Collectively, these limitations place metric extraction and exhaustive entity
detection outside the model's authority. Deterministic procedures instead
measure coordinates, resolve scale, and apply named rules to candidate entities.
Their repeatability does not eliminate missed or false detections when an office
convention falls outside the encoded grammar. Annotations also constitute an
open vocabulary that fixed parsing rules cannot fully cover. Unsupported
symbols can leave meaningful linework unassigned, and geometric anomalies alone
do not reveal the drafter's intent. Semantic review is therefore reserved for
residual cases that deterministic geometry can identify but cannot interpret.

The proposed architecture enforces this division of responsibilities. Geometry
supplies coordinates, repair candidates, and the initial entity set, while the
vision-language model can act only through a closed schema of typed operations.
Additions require supporting axis or glyph linework; other operations use
class-specific caps, range checks, and judging because no single geometric
predicate can verify every semantic change. Proposals that fail these checks
are rejected, while model-call or parsing failures preserve the input layout.
A semantically incorrect
proposal can still pass the available checks, so the framework produces a
reviewable model draft rather than a certified structural model.

\subsection{System architecture}
\label{sec:overview}

\begin{figure}[t]
\centering
\includegraphics[width=\linewidth]{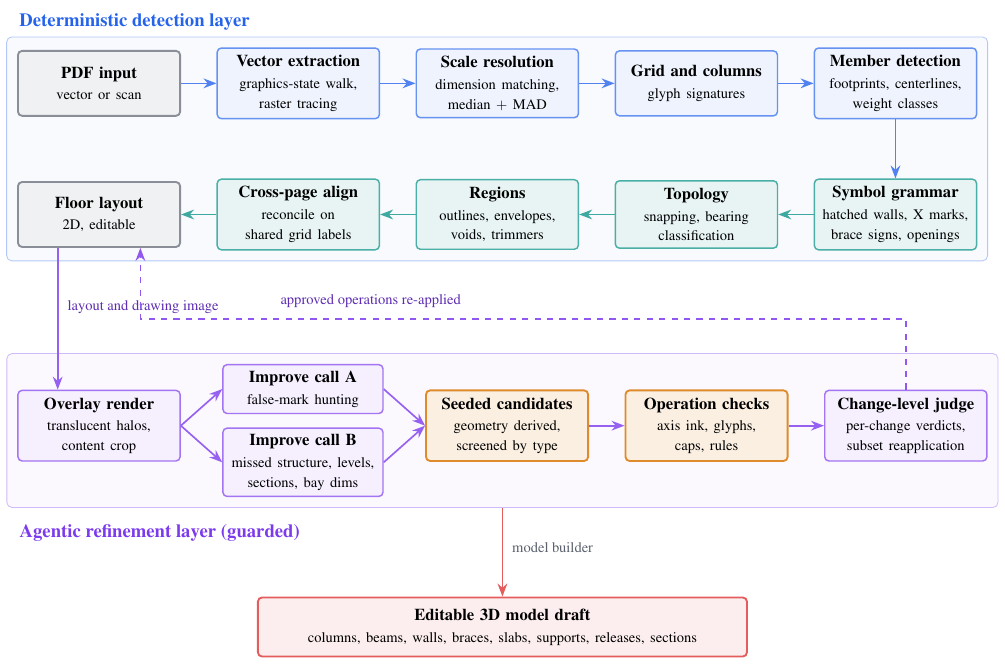}
\caption{System architecture with the deterministic detection layer, the
guarded agentic refinement layer, and the model builder.}
\label{fig:pipeline}
\end{figure}

The proposed system converts a structural framing-plan PDF into an editable finite-element
model draft through three stages (Figure~\ref{fig:pipeline}). The input may be
a vector export from CAD or BIM software, or it may be a scanned drawing. First,
deterministic extraction produces a typed two-dimensional layout containing
grids, columns, beams, walls, braces, slab regions, and openings in model-space
meters. The implementation applies the named rules in a fixed order, and the
resulting layout remains editable. Second, the guarded agentic layer reviews an
overlay of this layout on the source drawing
and proposes corrections through a closed operation schema. Only operations
that satisfy the applicable guards and judging procedure are accepted. This
admission rule applies to typed entity edits; printed-level metadata follows the
separate validation path described later in this section.

The third stage converts the refined floor layouts and user-specified story
heights into a three-dimensional model draft. The model builder places beams at
floor elevations, erects columns between stories, instantiates walls and braces,
and meshes slabs around detected openings. Parsed designations determine
sections and materials when they are available, while configured defaults fill
unresolved properties. The builder then assigns base restraints and
secondary-member releases. Because the intermediate layout remains editable,
every accepted detection or repair can be inspected and corrected before
analysis.

Two design principles govern the information flow. First, scale,
collinearity, coverage, bearing, and containment are computed by explicit
procedures rather than predicted by a task-specific detector, although their
thresholds remain design choices. Second, the language model is consulted only
after the initial layout exists, and its answers remain proposals until they
pass the available checks. The following subsections specify the deterministic
and agentic stages. Section~\ref{sec:results} evaluates the complete vector-PDF
pathway, including guarded refinement, while
Section~\ref{sec:rasterillustration} provides a non-evaluative illustration of
the raster and model-assembly pathways.

\subsection{Deterministic detection layer}
\label{sec:deterministic}

The deterministic layer converts drawing evidence into editable model geometry
through three dependent phases. Primitive extraction creates a common vector
representation and resolves the drawing scale. Ordered structural-element
passes then detect grids, columns, members, and symbols before constructing the
bearing topology. Finally, cross-drawing reconciliation aligns related plans,
and the model builder assembles their entities into an editable
three-dimensional draft.

\subsubsection{Primitive extraction and scale resolution}
\label{sec:extract}

The extraction stage converts either vector or image-only PDF input into a
common set of paper-space geometric primitives. For vector exports, a custom
parser traverses PDF graphics operators while tracking the
transformation stack, line width, dash pattern, stroke and fill colors, and
path-painting operators. Its output comprises attributed segments, closed
polygons, arcs recovered from flattened B\'ezier runs by least-squares circle
fitting, and positioned text runs with their rotations. For image-only PDFs,
the adaptive thresholding method of \citet{sauvola2000adaptive} binarizes the
embedded bitmap, which is then traced into stroke primitives. Both paths feed
the same geometric detectors, although only the vector path preserves a
machine-readable text layer. Retaining geometric primitives, rather than pixels
alone, also preserves vector endpoints and line weights, an advantage shared with
vector-domain symbol-spotting methods
\citep{fan2021floorplancad,yang2023vectorfloorseg}.

\paragraph{Scale resolution by dimension consensus.}
\label{sec:scale}

All downstream geometric tolerances are expressed in physical units, so the
drawing coordinates must first be converted from paper millimeters to model
millimeters. An incorrect conversion factor affects every modeled coordinate.
The scale resolver therefore precedes entity detection and reports both its
method and confidence (Algorithm~\ref{alg:scale}). Each dimension candidate pairs a
printed distance, such as 6000~mm, with the line that represents that measured
span. The matcher uses strict geometric conditions to reduce incorrect pairings:
the text must sit within 12~mm of a thin line whose direction
agrees with the text rotation within $6^{\circ}$, must project into the
middle 90\% of that line. Both line ends must also carry a terminator: either a
tick stroke of length 0.8 to 6~mm whose midpoint lies within
1.2~mm of the end at a relative angle above $20^{\circ}$, or a filled
arrowhead centroid within 1.6~mm.

Each surviving candidate $i$ yields a scale estimate $s_i = D_i / d_i$,
where $D_i$ is the stated distance parsed under the drawing's unit system
and $d_i$ is the measured line length in paper millimeters. Consensus is
obtained by rejecting outliers using the median absolute deviation and
averaging the remaining estimates,
\begin{equation}
\mathcal{I} = \bigl\{\, i : \lvert s_i - \tilde{s} \rvert \le
\tau \,\bigr\},\qquad
\tau = \max\bigl(0.01\,\tilde{s},\ 4\,\mathrm{MAD}\bigr),\qquad
\hat{s} = \frac{1}{\lvert \mathcal{I} \rvert}
\sum_{i \in \mathcal{I}} s_i,
\label{eq:scale}
\end{equation}
with $\tilde{s} = \operatorname{median}\{s_i\}$ and median absolute deviation
$\mathrm{MAD} = \operatorname{median}\{\lvert s_i - \tilde{s}\rvert\}$.
The threshold includes a minimum band equal to 1\% of the median. This band
admits small measurement perturbations when the MAD is zero or nearly zero,
although exactly equal candidates would also survive a zero-width band.
Confidence is defined as the inlier fraction $\lvert \mathcal{I} \rvert / n$.
When $\hat{s}$ falls within
1.5\% of one of the fifteen configured reference scales it snaps to that
scale, which removes residual measurement noise on conforming drawings.
A valid two-click calibration takes precedence over dimension consensus. When
neither source is available, a drawing with no usable dimension receives an
assumed scale of 1:100 with zero confidence, which specifically triggers review
by the agentic layer.

\begin{algorithm}[t]
\SetAlgoLined
\DontPrintSemicolon
\caption{Resolution of drawing scale by dimension consensus.}
\label{alg:scale}
\KwIn{primitives $G$ (paper mm), unit system $u$, optional two-point
calibration $c=(a,b,D)$}
\KwOut{scale $\hat{s}$ (model mm per paper mm), method, confidence}
\KwCfg{text offset $12$; text param $[0.05, 0.95]$; angle $6^{\circ}$;
tick length $[0.8, 6]$, midpoint $1.2$, angle $>20^{\circ}$; arrowhead
$1.6$; min line $3$ (all paper mm); configured reference scales
$\mathcal{S}_0=\{10,20,25,30,40,50,75,100,125,150,200,250,300,400,500\}$}
\If{$c$ is valid and $\lvert a-b\rvert>0.5$ mm}{
  $\hat{s}\leftarrow D/\lvert a-b\rvert$; snap to $\mathcal{S}_0$ within
  1.5\%; \Return $(\hat{s},\ \textnormal{user-calibrated},\ 1)$;
}
$T \leftarrow \{s \in G.\text{segs} : \neg s.\text{dashed},\ s.w \le 0.3,\ \lvert s \rvert \ge 3\}$
  \tcp*{thin bucket}
$S \leftarrow [\,]$\;
\ForEach{text run $t \in G.\text{texts}$}{
  $D \leftarrow \textsc{ParseDimension}(t, u)$
    \tcp*{$24'\text{-}6''$, $7500$, $6.40$; reject $D < 100$ mm}
  \lIf{$D = \varnothing$}{\textbf{continue}}
  $\ell \leftarrow \arg\min_{s \in T} \lvert \mathrm{off}(t, s) \rvert$
    subject to the configured offset, projection, angle, and two-terminator tests\;
  \lIf{$\ell \neq \varnothing$}{append $D / \lvert \ell \rvert$ to $S$}
}
\lIf{$S = \varnothing$}{\Return $(100,\ \textnormal{assumed},\ 0)$
  \tcp*[f]{return assumed scale and warning}}
compute $\tilde{s}$, $\mathrm{MAD}$, $\tau$, $\mathcal{I}$, $\hat{s}$ by
Eq.~\eqref{eq:scale}\;
\lIf{$\exists\, \sigma \in \mathcal{S}_0 : \lvert \hat{s} - \sigma \rvert / \sigma \le 0.015$}{$\hat{s} \leftarrow \sigma$}
\Return $(\hat{s},\ \textnormal{dimension-consensus},\ \lvert \mathcal{I} \rvert / \lvert S \rvert)$\;
\end{algorithm}

\subsubsection{Structural-element detection, symbol rules, and topology}
\label{sec:members}

Structural detection follows an ordered sequence in which each pass constrains
the next. Grid recovery establishes the reference lattice, column glyphs define
support locations, member and symbol rules classify intervening linework, and
the topology pass connects retained entities. The ordering also prevents
annotations and structural symbols from being incorporated into members during
line chaining.

A grid line is generally a chain of dash-dot strokes, identified by a dash
array of at least four entries and anchored by a bubble at one or both ends.
A long solid line attached to a labeled bubble is also admitted when an
exporter omits dash patterns. A bubble
is a circle of radius 3 to 7.5~mm containing a one- or two-character
label whose center lies within 80\% of the radius. Some exporters represent
circles as tessellated polygons, which are recovered using a polygon-roundness
test. Label content, rather than line direction, assigns the grid family. This
criterion separates letters and digits correctly on a rotated wing, where an
orientation-based classification would fail. A circular mean over each family
serves only to check that the two families are approximately perpendicular, and
bay distances are then derived from the intersection lattice.

The column detector converts compact structural glyphs into locations,
footprint dimensions, and orientations but does not assign a material. A
candidate is a closed glyph whose principal extents lie between 60 and
2000~mm on the long axis and between 30 and 2000~mm on the short axis, with
an aspect ratio no greater than 6 so that member end caps do not qualify. The
detector samples the interior width at five stations along the principal axis.
Wide ends with a pinched middle indicate a wide-flange shape, wide ends with a
wide middle indicate a rectangle, and a single wide end indicates a tee. A
concentric white knockout identifies a hollow glyph. Material and section
assignment is deferred to model assembly because the glyph alone does not
provide sufficient semantic evidence.

Two fallback rules accommodate common differences among exports. When a drawing
contains
no filled structural polygons, the detector also tests unfilled glyphs and
closed loops assembled from loose strokes. A stroked circle is admitted only
when at least two hatch strokes lie inside it; this condition suppresses north
arrows and detail marks in the same size range. The estimated orientation is folded to
$[0^{\circ},180^{\circ})$. Values within $10^{\circ}$ of $90^{\circ}$ are retained
as $90^{\circ}$, while all other values default to $0^{\circ}$ because orientation
estimates for nearly square glyphs are unstable.

Member detection then converts elongated primitives into beam and wall
candidates (Algorithm~\ref{alg:members}). Within the supported grammar, beam
footprints are closed, unfilled polygons drawn at their true width, whereas beam
centerlines are chains of collinear medium-weight strokes. These two
representations cover the concrete, glulam, and steel conventions encoded by the
benchmark generator without inferring material from linework alone. Stroke
weights are separated into hatch, medium, and heavy classes using percentiles
of each drawing's width histogram together with fixed implementation intervals.
This combination reduces sensitivity to different pen tables. Walls are
recognized before beams
because a wall footprint can otherwise be interpreted as a wide beam. When
several coincident outlines describe one wall, the thinnest pair defines its
centerline.

The topology pass then converts geometry into structural relations. Fragments are chained
across gaps up to 450~mm. Endpoints first seek a column within half its
maximum glyph extent plus 600~mm, provided the member axis passes within
that extent plus 120~mm. Unsnapped ends then seek a wall within half its
thickness plus 600~mm and, finally, a beam within half its drawn width
plus 600~mm (150~mm is used when width is unavailable). Let
$A_q(p,e)$ state that endpoint $p$ of member $e$ satisfies the corresponding
snap test for support type $q$, and let $\operatorname{first}$ return the
first qualifying type in the ordered tuple $(\mathrm{column},\mathrm{wall},
\mathrm{beam})$, or $\mathrm{free}$ when none qualifies. The resulting bearing
and classification predicates are
\begin{equation}
\beta_e(p)=\operatorname{first}_{q\in(\mathrm{column},\mathrm{wall},\mathrm{beam})}
\{q:A_q(p,e)\},\qquad
K(e)=
\begin{cases}
\mathrm{reject}, & \beta_e(a)=\beta_e(b)=\mathrm{free},\\
\mathrm{secondary}, & \mathrm{beam}\in\{\beta_e(a),\beta_e(b)\},\\
\mathrm{girder}, & \text{otherwise}.
\end{cases}
\label{eq:bearing}
\end{equation}
Spans are split at interior columns and provisional column-to-column girders
before final labels are assigned. Rejecting a member with two unsupported
endpoints removes legend strokes and title-block rules without relying on a
fixed crop defined by page location.

After topology construction, the symbol grammar distinguishes structural
symbols from member linework. Within the supported drawing conventions, shear walls are represented as
hatched bands between columns, framed openings as X-marked rectangles, and
vertical braces as X or V signs on bay edges. The detector encodes these
representations as explicit geometric signatures rather than learned
appearance. Table~\ref{tab:rules} states the eleven rules;
Algorithm~\ref{alg:members} places the ten member and topology rules in
execution order, while leader-dot rejection belongs to the preceding column
pass. These rules encode conventions represented in the development corpus,
and their thresholds and signatures were derived using only the \pdbench{}
development half. No result from the held-out \pdtest{} half informed a
revision. The rules should therefore be understood as an explicit and testable
grammar rather than as universal drafting conventions.

Two properties support evaluation across drawing scales. First, each threshold
uses the units appropriate to the mechanism that generates the relevant mark.
Symbol tests use paper millimeters because drafters select printed symbol sizes
for legibility, independent of building scale. Structural tests use model-space
meters because physical bay dimensions remain invariant to drawing scale. A
single unit system for both classes would make one class of thresholds dependent
on drawing scale. Second, the rules consume symbols before interpreting the
remaining geometry: X marks are removed before chaining, dimension lines are
excluded before member formation, and walls are assigned before beams. Without
this ordering, residual symbol strokes can be classified as false members.
Table~\ref{tab:detcfg} consolidates the configured units, thresholds, and roles.

\begin{algorithm}[t]
\SetAlgoLined
\DontPrintSemicolon
\caption{Member and symbol detection with topology construction.}
\label{alg:members}
\KwIn{primitives $G$, scale $\hat{s}$, columns $C$, matched dimension
lines $\Delta$}
\KwOut{beams, walls, braces, openings, slab regions}
\KwCfg{paper mm: hatch corridor $6$, X-mark $[4,22]$, midpoint $2.5$,
dedup offset $\max(2, 0.75w)$; model mm: member $\ge 450$, wall
$[80, 650]$, opening side $\ge 1000$; meters: wall slop $0.3$, extent
guard $3$}
\tcp{Stroke classes: percentiles of this drawing's width histogram,
unioned with native windows}
$(h, m_{\mathrm{lo}}, m_{\mathrm{hi}}) \leftarrow \textsc{WeightClasses}(G)$\;
\tcp{R1 walls before beams: a wall footprint reads as a wide beam}
$W \leftarrow$ closed unfilled polys, weight in $[m_{\mathrm{lo}}, m_{\mathrm{hi}}]$, length
$\ge 600$ mm, width $\in [80, 650]$ mm, elongation $\ge 2.5$, containing
$\ge 3$ hatch midpoints\;
group coincident candidates (lateral $0.75t$, overlap $>0.6$); keep the
thinnest as the centerline\;
\tcp{R1b hatched bands between adjacent columns, longest first}
\ForEach{axis-aligned adjacent column pair with paper span $L_p$ and
$L_p\hat{s}\in[1500,20000]$ mm}{
  $\Theta \leftarrow$ strokes of length $1.2$--$30$ mm at
    $15^{\circ}$--$75^{\circ}$ to the drawing axes, midpoint within the
    $\pm 6$ mm corridor, and axial parameter $t \in (0.06, 0.94)$\;
  \lIf{$\lvert \Theta \rvert < \max(6, L_p/(8\,\mathrm{mm}))$}{\textbf{continue}}
  sort offsets $0\le d_{(1)}\le\cdots\le d_{(n)}$ and set
  $t_{\text{raw}} \leftarrow 2d_{(\lfloor0.9n\rfloor+1)}\hat{s}$
    \tcp*{R1c gate the RAW value, before any clamp}
  \lIf{$t_{\text{raw}} < 100$ mm}{\textbf{continue}
    \tcp*[f]{else corner tails become false walls}}
  $t_w\leftarrow 5\,\mathrm{round}(\min(600,t_{\text{raw}})/5)$ mm\;
  accept unless $\ge 5$ of $9$ samples are already covered by an accepted wall\;
}
\tcp{R2--R3 symbols are consumed before geometry}
$\mathcal{X} \leftarrow$ short diagonal pairs ($4$--$22$ mm, shared
midpoint $\le 2.5$ mm, opposite slopes) \tcp*{erased: brace signs}
$\mathcal{O} \leftarrow$ long diagonal pairs with a shared midpoint whose
four endpoints form a rectangle with sides $\ge 1$ m and $\ge 6$ of $8$
inked samples on all four edges \tcp*{framed openings}
\tcp{R5 members: two representations, one topology}
$F \leftarrow$ closed unfilled polys, $4$--$8$ vertices, length $\ge 250$
mm, width $\in [30, 800]$ mm, aspect $\ge 2$ \tcp*{footprints}
$P \leftarrow \{s : \neg\text{dashed},\ s \notin \mathcal{X},\ w \in
[m_{\mathrm{lo}}, m_{\mathrm{hi}}],\ \lvert s \rvert \hat{s} \ge 450,\ s \notin \Delta\}$
  \tcp*{R5 excludes dimension lines}
$K \leftarrow \textsc{ChainCollinear}(P, 0.75^{\circ}, 0.3, \text{gap})$\;
\ForEach{$k \in K$}{
  \lIf{$\exists f \in F$ with relative angle $\le 12^{\circ}$ \textnormal{and} midpoint offset
    $< \max(2, 0.75 w_f)$}{discard \tcp*[f]{R4: the angle gate is
    required}}
}
$\textsc{Topology}$: chain fragments, snap ends to columns and walls,
split at interior columns and girder crossings, assign bearings\;
\lForEach{member}{reject if both ends bear on nothing \tcp*[f]{R10}}
\tcp{R6--R9 post-conditions}
delete members whose midpoint lies inside an opening (R6); add perimeter
trimmers where $\le 1$ of $5$ samples is framed (R7); delete beams
covered on $\ge 3$ of $5$ samples within $t/2 + 0.3$ m of a wall (R8);
delete members outside the column bounding box inflated by 3 m (R9)\;
\end{algorithm}

\begin{table}[t]
\centering
\caption{Eleven detection rules in the supported drafting grammar.}
\label{tab:rules}
\small
\begin{tabular}{@{}p{0.26\linewidth}p{0.68\linewidth}@{}}
\toprule
Rule & Statement \\
\midrule
Hatched-band wall & The rule classifies a qualifying diagonal-stroke corridor between adjacent columns as a shear wall; thickness is twice $d_{(\lfloor0.9n\rfloor+1)}$, and the raw value must reach 100~mm before capping and rounding. \\
X-in-rectangle opening & The rule classifies two long diagonals with coincident midpoints whose endpoints form an inked rectangle as a framed floor opening. \\
X-mark eraser & Under the supported grammar, two short diagonals (4--22~mm on paper) crossing at a shared midpoint are treated as a symbol and removed before member chaining. \\
Angle-gated dedup & The duplicate test for centerlines measures offset to an infinite line and therefore requires parallelism within $12^{\circ}$; without the gate, joists whose midpoints lie on a perpendicular girder's line are deleted. \\
Dimension-line exclusion & Segments riding a dimension line already matched by the scale resolver are treated as annotation rather than members. \\
Void rule & Candidates whose midpoints fall strictly inside a detected opening region are removed as opening-symbol remnants. \\
Trimmer completion & A geometry-screened opening receives perimeter trimmer beams where its edges are uncovered. \\
Wall-coverage deletion & A beam whose sampled midline lies within half the wall thickness plus 0.3~m of a wall centerline duplicates that wall and is removed. \\
Column-extent guard & Candidates whose midpoints fall outside the column bounding box inflated by 3~m are classified as drawing furniture. \\
Bearing invariant & The implementation assumes a retained floor member bears on a column, wall, or beam at one or both ends; candidates that bear on neither end are rejected. \\
Leader-dot rejection & A glyph is discarded as annotation when it is both far smaller than the drawing's median column and has a thin annotation line terminating inside it. Neither test is applied alone: column sizes legitimately vary, and a beam end also terminates at a column. \\
\bottomrule
\end{tabular}
\end{table}

\begin{table}[t]
\centering
\caption{Deterministic-layer configuration. Paper millimeters govern tests of
printed symbols, whereas model-space millimeters and meters govern structural
tests.}
\label{tab:detcfg}
\small
\setlength{\tabcolsep}{4pt}
\begin{tabular}{@{}lp{0.40\linewidth}p{0.33\linewidth}@{}}
\toprule
Constant & Value & Role \\
\midrule
\multicolumn{3}{@{}l}{\textit{Scale resolution}} \\
Text-to-line offset, angle & 12 mm paper, $6^{\circ}$ & Pairing a dimension string with its line \\
Text parameter window & $[0.05, 0.95]$ & Text must project into the line's middle \\
Terminator & tick $[0.8, 6]$ mm at $>20^{\circ}$, or arrowhead within 1.6 mm & Both ends required \\
Outlier band & $\max(0.01\tilde{s},\ 4\,\mathrm{MAD})$ & Minimum band for near-zero MAD \\
Reference-scale snap & $\le 1.5\%$ relative & 15 configured scales \\
\midrule
\multicolumn{3}{@{}l}{\textit{Grids and columns}} \\
Grid-line criterion & $\ge 4$ dash-array entries, or long solid line at labeled bubble & Center, phantom, and exporter fallback \\
Bubble radius, label & 3--7.5 mm paper, 1--2 chars within $0.8r$ & Family from label content \\
Column size window & 60--2000 mm long, 30--2000 mm short & Model-space glyph size \\
Column aspect & $\le 6$ & Excludes member end caps \\
Profile stations, signature & 5 stations, 25 samples & Ends $>0.75$, middle $<0.5$ gives wide flange \\
Rotation rule & within $10^{\circ}$ of $90^{\circ}$: retain $90^{\circ}$; otherwise $0^{\circ}$ & Principal-component noise otherwise reads as a diamond \\
Duplicate glyph & $0.6\max(w, d)$ & Suppresses the story-above copy \\
\midrule
\multicolumn{3}{@{}l}{\textit{Members}} \\
Weight classes & $\begin{aligned}h&=\mathrm{clip}(P_{20},0.10,0.22)\\m_{\mathrm{lo}}&=\min(0.40,\max(h+0.02,0.8P_{45}))\\m_{\mathrm{hi}}&=\max(m_{\mathrm{lo}}+0.20,\min(1.20,1.15P_{97}))\end{aligned}$ & Paper-mm histogram, unioned with native windows \\
Footprint & 4--8 vertices, $\ge 250$ mm, 30--800 mm, aspect $\ge 2$ & Concrete and glulam convention \\
Centerline chain & $0.75^{\circ}$, 0.3 mm offset, $\ge 450$ mm & Steel and alternate-export convention \\
Dedup angle gate & $12^{\circ}$, offset $\max(2,\ 0.75w)$ mm paper & Without the gate, joists are deleted \\
Brace chain & dashed, $\ge 800$ mm, $\ge 2$ X strokes within 6 mm & Confirmed and unconfirmed both kept \\
\midrule
\multicolumn{3}{@{}l}{\textit{Symbols}} \\
Wall hatch & 1.2--30 mm strokes; $15^{\circ}$--$75^{\circ}$; $\pm 6$ mm corridor; $n\ge\max(6,L_p/(8\,\mathrm{mm}))$ & Thickness $=2d_{(\lfloor0.9n\rfloor+1)}\hat{s}$, raw $\ge100$ mm, cap 600 mm \\
X-mark eraser & 4--22 mm paper, midpoint $\le 2.5$ mm, opposite slopes & Consumed before chaining \\
Opening & diagonals $\ge 1.2$ m, sides $\ge 1$ m, 6 of 8 edge samples & All four edges required \\
X-sign brace & 4--30 mm strokes, $\ge 2$ near midspan, both slopes & Wall-owned spans vetoed \\
\midrule
\multicolumn{3}{@{}l}{\textit{Topology}} \\
Fragment chaining & gap 450 mm, offset 140 mm, $2.5^{\circ}$ & Before any bearing test \\
End snap & column half-extent $+600$ mm (axis aim $+120$ mm); wall or beam half-width $+600$ mm & Precedence column, wall, beam; beam default width 150 mm \\
Split at column & $t\in(0.02,0.98)$; support clearance half-extent $+600$ mm & Re-cuts chained runs at interior columns \\
Split at girder & member $t\in(0.03,0.97)$; support $u\in[-0.02,1.02]$; piece $\ge300$ mm & Column-to-column provisional girders remain uncut \\
Wall-coverage deletion & $t/2 + 0.3$ m on $\ge 3$ of 5 samples & Removes coincident beam duplicates \\
Extent guard & column bbox $+\,3$ m & Preserves cantilever candidates \\
\bottomrule
\end{tabular}
\end{table}

\subsubsection{Reconciliation and model assembly}
\label{sec:assembly}

The final deterministic stage forms slab regions, aligns related drawings, and
assembles their structural entities into a model draft. Slab regions are
derived from three sources rather than by traversing faces in a planar graph,
because such traversal produces one face per framing bay without distinguishing
a slab from a light well. A dashed, unfilled phantom outline that encloses at
least one square meter is accepted directly as a slab boundary. If no such
outline exists, the column envelope, expanded by half a glyph width, defines
the slab extent. Openings are obtained from the structural-symbol pass or, for
architectural input, from stair and shaft boxes clipped to the grid lattice.

Related drawings must share a common coordinate frame before their entities can
be reconciled. Each drawing initially places its origin at the first detected
grid intersection, which can misalign a setback story beginning at grid B with
a lower story beginning at grid A. Shared grid labels provide the primary basis
for registration: the translation is the median positional difference among
common labels, which limits the influence of a single relabeled line. When two
drawings share no grid labels, the procedure selects the translation supported
by the greatest number of column pairs and accepts it only if a majority of the
columns align with the reference. If neither source supports a translation, the
drawing retains its local origin and is flagged for review. After registration,
reconciliation forms a consensus column lattice, shares the resolved scale when
appropriate, and permits a wall detection to replace a coincident beam
detection.

The model builder converts the reconciled layouts into a three-dimensional
finite-element model draft. Columns extend between stories after continuity is
matched across floors; beams are placed at floor elevations; walls become area
elements with the detected thickness; and brace marks instantiate diagonal, X,
V, or chevron components. Slabs are meshed around openings, base restraints are
assigned at the lowest story, and secondary members receive end releases.
Printed designations provide sections and materials when available. Otherwise,
the implementation combines neighboring-member labels, glyph shape, footprint
matches against configured steel, concrete, and glulam section catalogs, and
configured defaults. All fallback assignments require engineering review.

\subsubsection{Illustrative raster pathway and model assembly}
\label{sec:rasterillustration}

\begin{figure}[t]
\centering
\includegraphics[width=\linewidth]{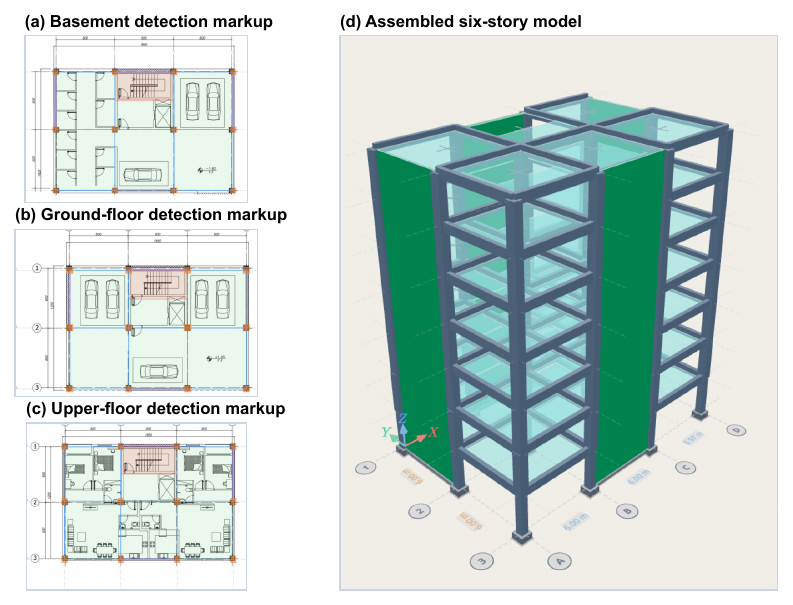}
\caption{Illustrative scanned-plan layouts with assumed lattice framing in
panels (a)--(c) and an independent six-story model draft in panel (d).}
\label{fig:realworld}
\end{figure}

Figure~\ref{fig:realworld} illustrates the raster and model-assembly pathways
using three scanned architectural plans from a six-story building drawing set.
Panels (a)--(c) show user-calibrated layouts processed through the raster pathway
described in Section~\ref{sec:extract}. The illustration uses a prescribed
story height of 3.2~m, with the first plan assigned to the lowest level, the
second repeated across four intermediate levels, and the third assigned to the
top level. This story mapping was supplied to the model builder rather than
inferred from the scans. Because raster processing produced no machine-readable
text, a 6.0~m consensus bay read from the drawing was supplied externally for
scale calibration. The input plans contained no explicit beam linework. After
detecting 12 column locations on each distinct plan, the workflow placed framing
members between adjacent lattice positions as a planning assumption requiring
engineering review. Panel (d) is an independent rendering of the intended
multistory model representation and is not an end-to-end output from the exact
layouts in panels (a)--(c). Because traceable entity-level annotations are
unavailable for this case, the figure is excluded from quantitative evaluation
and does not establish detection accuracy, model correctness, or
out-of-distribution generalization.

\subsection{Guarded agentic refinement}
\label{sec:agentic}

The deterministic layer can retain inconsistencies that geometry alone cannot
resolve. Examples include unexplained linework that may represent a missed
member, a detection placed over blank paper, a column mark displaced from its
glyph, or a drawing with an assumed scale. Here, \emph{agentic refinement}
denotes an iterative vision-language review that receives geometry-derived
candidates, proposes typed edits, and evaluates the resulting changes through a
separate review call. Its input is the \emph{layout}, defined as the typed set of
detected structural entities and their model-space geometry. Resolving each
inconsistency requires a judgment about drawing intent. A vision-capable
language model can propose an interpretation, but generative models can produce
unsupported content
\citep{ji2023hallucination}. Current vision-language models also show
limitations on low-level visual tasks \citep{rahmanzadehgervi2024blind}.
Consequently, the refinement layer treats the model as a reviewer with no
direct authority to modify the layout.

\subsubsection{Refinement workflow and permitted edits}
\label{sec:refineloop}

Algorithm~\ref{alg:refine} defines the refinement workflow and prevents the
model from modifying the layout directly. A \emph{typed operation} is one of
twelve named changes accompanied by its required fields, whereas a \emph{seed}
is an operation proposed by a deterministic candidate generator rather than by
the model. The \textsc{ValidateAndApply} routine checks required fields and
operation-specific evidence, applies the surviving edits to a copy of the input
layout, and records both accepted and rejected operations. No other procedure
can create a revised layout. This interface restricts the action space but does
not provide geometric proof for every accepted action. Addition guards test
drawing linework, whereas attribute changes and most deletions rely on typed
fields, class-specific limits, and subsequent judging. If no candidate survives,
or if neither the complete batch nor a reviewed subset is accepted, the routine
returns the original entity layout. This \emph{fail-closed} behavior applies to
entity edits but not to the separately reported page-level metadata described
below.

\begin{algorithm}[t]
\SetAlgoLined
\DontPrintSemicolon
\caption{Guarded agentic refinement of one drawing.}
\label{alg:refine}
\KwIn{deterministic layout $L$, drawing primitives $G$, page raster $R$}
\KwOut{refined layout, or $L$ unchanged}
\KwCfg{$\alpha{=}0.55$, halo $2.8\times$, crop $42$\,mm, $1600$\,px;
provider defaults; high resolution; $8192$ tokens; $\le12$ ops/call,
$\le24$ merged}
$O_{\text{img}} \leftarrow \textsc{Overlay}(L, R, \alpha, \text{halo}, \text{crop})$\;
\lIf{$O_{\text{img}} = \varnothing$}{\Return $L$ \tcp*[f]{no vision call is made}}
$D \leftarrow \textsc{Digest}(L)$ \tcp*{entities, dimensions, per-bay joist counts}
\For{$\text{attempt} \leftarrow 1$ \KwTo $2$}{
  $(\Omega_{-}, \Omega_{+}) \leftarrow$ \textbf{parallel}
    $\textsc{Vision}(O_{\text{img}}, D, \textsc{FocusFalse})$,
    $\textsc{Vision}(O_{\text{img}}, D, \textsc{FocusMissed} \parallel \Sigma_{\text{scan}})$\;
  $s \leftarrow \textsc{CalibrationSeed}(L)$\;
  $\Sigma \leftarrow \textsc{InklessMarks}(L,G) \cup \textsc{UnmarkedColumns}(L,G) \cup
    \textsc{UnmarkedSpans}(L,G) \cup \textsc{UnmarkedJoists}(L,G)$\;
  $(\Omega, X_c) \leftarrow \textsc{Canonicalize}(\Omega_{-} \frown \Omega_{+} \frown \Sigma)$
    \tcp*{deduplicate effects; drop conflicts}
  \If{$s \neq \varnothing$ \textbf{or} $\Omega$ contains calibration}{
    $\Omega \leftarrow [\textsc{SelectCalibration}(s,\Omega)]$
      \tcp*{exclusive transaction}
  }
  $\Omega \leftarrow \textsc{Truncate}(\Omega,\ 24)$\;
  \lIf{$\Omega = \varnothing$}{\Return $L$}
  $(L', A, \Omega_A, X) \leftarrow \textsc{ValidateAndApply}(L, \Omega, G)$
    \tcp*{$A$: change list, $X$: rejections}
  \lIf{$A \neq \varnothing$}{\textbf{break}}
  feed $X$ back as corrective context\;
}
\lIf{$A = \varnothing$}{\Return $L$}
$v \leftarrow \textsc{ValidateJudge}(\textsc{Judge}(O_{\text{img}}, \textsc{Overlay}(L'), A), |A|)$\;
\lIf{$v = \varnothing$}{\Return $L$}
\lIf{$v.\text{verdict} = \textnormal{accept}$}{\Return $L'$}
$B \leftarrow v.\text{bad\_changes}$ \tcp*{unique, in-range integer indices}
\If{$0 < |B| < |A|$}{
  $(L'', A'') \leftarrow \textsc{ValidateAndApply}(L, \{\omega_i \in \Omega_A : i \notin B\}, G)$
    \tcp*{salvage}
  \If{$|A''| = |A|-|B|$}{
    $v'' \leftarrow \textsc{ValidateJudge}(\textsc{Judge}(O_{\text{img}}, \textsc{Overlay}(L''), A''), |A''|)$\;
    \lIf{$v'' \neq \varnothing$ \textbf{and} $v''.\text{verdict} = \textnormal{accept}$}{\Return $L''$}
  }
}
\Return $L$\;
\end{algorithm}

Two focused calls separate the review objectives. The false-mark pass deletes
detections or moves columns, whereas the missed-structure pass adds members,
sets attributes, or calibrates scale. This decomposition separates destructive
and constructive proposals before they enter a common validation path. The
standard workflow uses two proposal calls and one judging call; retries and
subset review can increase the total to six logical model invocations, excluding
up to three transport attempts per invocation. The \emph{judge} is a separate
vision-language call that reviews the numbered accepted changes, and
\emph{salvage} denotes reapplication of only the changes that it did not reject.
Section~\ref{sec:prompts} reproduces the shared overlay legend, the improve
system prompt, the two focus suffixes, and the judge prompt.

The operation vocabulary defines the permitted interface between the model and
the layout (Schema~\ref{box:ops}). Its twelve operation names cover deletion,
attribute assignment, column movement, structural additions, and two forms of
scale calibration. Coordinates are expressed in model-space meters with the
$y$-axis directed upward. Target identifiers must match entities in the layout
\emph{digest}, which is a compact list of current entities, dimensions, and
per-bay member counts supplied to the model. The provider schema requires an
operation name, after which \textsc{ValidateAndApply} verifies the fields
required for that operation and discards malformed proposals. The prompt also
requests a free-text reason for review by the judge, although the machine schema
treats this field as optional. Because generator-written provenance and
model-written reasons are not stored in separate trusted fields, provenance
text cannot constitute independent evidence. Section~\ref{sec:conclusions}
revisits this limitation.

\begin{schemabox}[label={box:ops}]{The typed operation vocabulary, with the
literal examples embedded in the prompt}
\begin{lstlisting}[style=json]
§§// destructive and attribute ops: id and kind copied verbatim from the digest§§
{"op":"delete","kind":"beam|column|wall|brace|region","id":"beam_12","reason":"..."}
{"op":"set_section","kind":"beam","id":"beam_3","sectionName":"W360X134"}
{"op":"set_wall_thickness","id":"wall_2","thicknessMm":250}
{"op":"set_region_kind","id":"region_5","regionKind":"slab|opening"}
{"op":"set_beam_kind","id":"beam_7","beamKind":"girder|secondary"}
§§// slab thickness read from a keynote; no id applies it to every floor region§§
{"op":"set_region_thickness","thicknessMm":200,"reason":"keynote 1 reads 200 mm slab"}

§§// constructive ops: coordinates snap to columns and grid intersections§§
{"op":"add_column","x":0,"y":14.6,"reason":"glyph at grid A-3 carries no mark"}
{"op":"move_column","id":"col_8","x":0,"y":14.6,"reason":"marked 0.6 m off the glyph"}
{"op":"add_beam","x1":0,"y1":7.2,"x2":6.4,"y2":7.2,"beamKind":"secondary",
 "reason":"drawn line between girder_3 and girder_4 has no overlay"}
{"op":"add_wall","x1":0,"y1":0,"x2":7.5,"y2":0,"thicknessMm":300}

§§// metric ops: range/state checked; seeded values derive from parsed dimensions§§
{"op":"calibrate_scale","gridA":"A","gridB":"E","realMm":25600,
 "reason":"dimension strings state 4 x 6400 between A and E"}
§§// grid-free calibration: one value per gap between consecutive column lines§§
{"op":"calibrate_bays","axis":"x","bayMm":[6000,6000,6000],
 "reason":"no grids detected; the top chain reads 600|600|600 in cm"}
\end{lstlisting}
\end{schemabox}

The two calibration operations address different forms of available evidence.
In a vector PDF, positioned text runs expose dimension strings, section
designations, grid labels, and level marks to deterministic parsing. An
image-only PDF lacks this machine-readable text layer, so these items require
visual interpretation. The \texttt{calibrate\_scale} operation uses two
detected grid labels and their stated distance and is therefore unavailable
when raster processing does not recover a labeled grid pair. The
\texttt{calibrate\_bays} operation instead uses the complete bay-dimension chain
along one axis. It clusters detected column centers into lines, requires one
stated bay length for every consecutive gap, and computes the scale factor from
the stated total and measured extent. Its guard checks completeness, factor range,
and internal consistency, but it does not independently read the dimension
text. A plausible but incorrectly transcribed visual value may therefore pass.

The layer also reports printed floor-level metadata outside typed-operation
admission. Although these values do not alter detected entity geometry, they can
affect downstream story configuration. They bypass operation judging and
receive only list-format and plausibility checks, so user confirmation remains
necessary. The output is an ascending set of printed floor elevations in
meters. The implementation interprets one elevation as a floor location and two
as adjacent levels that define a story height without implying repetition. It
interprets three or more regularly spaced elevations as a possible
repeated-story list. For vector input, at least three elevations must also be
left-aligned before they are treated as a list; otherwise, they remain
individual references. These interpretation rules are implementation
assumptions whose accuracy was not evaluated.

Slab thickness follows the typed-operation path. When the review identifies a
keynote or general note that applies to the floor plan, a page-wide attribute
operation proposes the parsed value for every floor region on that drawing. The
operation must pass the region-existence guard, the 50--1200~mm range guard, and
the judge. The assumed scope of the note is specific to the supported workflow
and remains subject to user review.

\subsubsection{Geometry-seeded candidates and rule-based guards}
\label{sec:seeds}

Five deterministic generators create repair candidates before operation
admission. \textsc{CalibrationSeed} proposes a scale factor from dimension
consensus; \textsc{InklessMarks} proposes the removal of entities without
supporting linework; \textsc{UnmarkedColumns} proposes columns at vacant grid
intersections containing glyph-like linework; \textsc{UnmarkedSpans} proposes
members between adjacent columns when their axis is drawn; and
\textsc{UnmarkedJoists} proposes missing members in repeated bay patterns. The
three constructive lists are also included in the model prompt as visual-review
context. Model proposals and the four non-calibration seed lists are
canonicalized before application of the 24-operation cap: duplicate effects are
reduced to one, and contradictory effects targeting the same mutation slot are
discarded. Calibration is excluded from ordinary batching. When either a
deterministic seed or a model proposal requests calibration, the transaction
contains only one calibration operation, with priority given to the
dimension-derived seed. Geometry and attribute edits are deferred until a new
pass uses the corrected scale. The generators screen geometry but do not
establish structural meaning; for example, a span with continuous linework may
still be a dimension or annotation line. Table~\ref{tab:agenticcfg}
states the implemented tests. An unmarked column needs glyph-shaped ink at
a free grid intersection; an unmarked span needs at least 0.60 axis
coverage and is typed by its diagonal-stroke signature; a joist candidate
must occupy a repeated bay fraction; inkless beam and wall marks fall below
0.25 axis coverage, while column marks use the glyph test; and a calibration
seed requires at least three dimension-derived factors, 60\% of which lie
within 2\% of their median.

\begin{promptbox}[label={box:inkscan}]{Geometry-derived repair candidates supplied to the missed-structure review}
\begin{lstlisting}[style=prompt]
@@DETERMINISTIC INK SCAN@@ (candidate locations screened against drawing vectors):
MISSING-COLUMN candidates (a drawn column glyph sits at these grid intersections with NO orange mark): 1. (0.00, 14.60)  2. (7.20, 14.60). Emit add_column with EXACTLY these coordinates for each real glyph.
MISSING-MEMBER spans between adjacent marked columns (drawn ink, no marked beam/wall): 1. (0.00, 0.00) to (7.50, 0.00) - HATCHED/FILLED band, classified as a wall candidate: emit add_wall only if the band is structural  2. (7.50, 0.00) to (14.40, 0.00) - plain line: emit add_beam only if the line is structural. Use EXACTLY these coordinates.
MISSING-JOIST spans (drawn linework matching the repeating joist pattern, no markup member): 1. (2.40, 0.00) to (2.40, 7.20). Emit add_beam with EXACTLY these coordinates for EACH, unless a span is clearly not a member (a dimension line or text underline).
\end{lstlisting}
\end{promptbox}

The emphatic prompt labels describe geometric stroke evidence, not proof of
structural meaning. Because this wording may bias both the proposing model and
the judge toward the seeded class, its effect requires a prompt ablation in
which candidate provenance is withheld.

Operation guards determine which proposed changes may reach the judge. Each
guard is a deterministic function of a copied layout and, for selected
additions, the extracted drawing primitives. Table~\ref{tab:agenticcfg} lists
the principal constants. For an added beam or wall with model endpoints $a,b$,
let $T$
map model coordinates to paper coordinates and let $\mathcal{S}$ be the
extracted strokes. Axis-ink coverage is
\begin{align}
q_i&=(1-\lambda_i)T(a)+\lambda_iT(b),
&\lambda_i&=\frac{i+1/2}{15},\\
C_\delta(a,b)&=\frac{1}{15}\sum_{i=0}^{14}
\mathbf{1}\!\left[\exists s\in\mathcal{S}: q_i\in
\operatorname{bbox}_{\delta}(s)\ \wedge\
d(q_i,s)\le\delta+\frac{w_s}{2}\right].
\label{eq:inkcoverage}
\end{align}
Here $w_s$ denotes stroke width, and $\operatorname{bbox}_{\delta}$ denotes the
stroke bounding box expanded by $\delta$. The admission guard uses
$\delta=2.4$~paper mm and requires $C_\delta\ge0.55$; because there are 15
samples, the smallest passing count is 9, corresponding to 0.60. This gate
applies to
\texttt{add\_beam} and \texttt{add\_wall}, not to moves, deletions, or
attribute operations. An added column instead requires glyph-shaped ink:
a nearby polygon centroid, a circle with a radius no greater than 12~paper mm,
or at least two strokes no longer than 12~paper mm within the stated search
radius. Long centerlines alone are insufficient.

The displaced-column guard distinguishes an incorrect mark from a misplaced
one. A deletion is refused if the mark has no glyph within a tight 3~mm
paper-space radius but an unoccupied grid intersection containing glyph-like
linework lies 0.3 to 1.5~m away in model space; in that case, the permitted
correction is a move. The asymmetric radii are deliberate. The guard uses 3~mm
at the mark because a correctly placed mark is centered on its glyph, and 6~mm
at the candidate because the search must accommodate the glyph dimensions. A
0.69~m displacement corresponds to 9~mm on paper at 1:75, so a model-space test
alone cannot distinguish the two cases.

For admitted spans, a diagonal-stroke classifier overrides the requested beam
or wall type. A stroke counts when its paper length is at
most 25~mm, its midpoint is within 5~mm of the span axis, and its
undirected angle relative to that axis is strictly between $15^{\circ}$
and $75^{\circ}$. For paper span length $L_p$, the span is classified as
hatched when
\begin{equation}
n_{\mathrm{diag}}\ge \max\!\left(4,\frac{L_p}{15~\mathrm{mm}}\right).
\label{eq:agenthatch}
\end{equation}
Both the four-stroke minimum and the length-dependent density condition must
therefore be satisfied. The test evaluates extracted strokes rather than filled
regions. An admitted span with qualifying hatching is typed as a wall; all other
admitted spans are typed as beams. An addition is refused when a beam, wall, or
brace
already occupies the same endpoints or covers at least half of the proposed
axis. The operation therefore cannot silently replace an existing member of a
different class.

The remaining operations use more limited checks. Deletions are capped by class;
column deletions also invoke the displaced-column refusal, and wall deletions
are refused whenever the hatch test passes. No general absence-of-linework
criterion applies to a model-proposed deletion. A column move is snapped,
restricted to 0.05--3~m, and rejected when its destination is occupied, but the
destination need not contain a detected glyph. Attribute operations validate
identifiers, enumerated values, numeric limits, and whether the requested value
would alter the layout. Section assignment includes an additional provenance
guard: a nonempty existing section cannot be overwritten, and an empty section
can be assigned only when the exact normalized designation occurs near the
entity in the local PDF text layer. Other attribute operations do not
independently establish evidence from the drawing. Calibration is permitted only
for an assumed or low-confidence scale and checks the grid pair, factor range,
and dead band. The guard does not independently reread a model-supplied
\texttt{realMm} value from raw dimension strings, whereas deterministically
seeded calibration derives that value from parsed dimensions.

\subsubsection{Change-level review and fail-closed behavior}
\label{sec:judge}

The judge conducts a second review after operation admission. It receives the
before and after overlays together with the authoritative, numbered change list,
and its prompt restricts assessment to the listed changes. A response is valid
only if it contains an exact verdict and a required \texttt{bad\_changes} array
of unique integer indices within the valid range. Acceptance requires an empty
array, whereas rejection requires at least one index; malformed or contradictory
responses fail closed. A proposed retained subset is then reapplied to the
original layout through the same guards. Salvage is permitted only when every
retained operation is reproduced exactly and a second strict judge response
accepts the replay. This procedure can remove a rejected addition without
implicitly accepting a partial or malformed transaction. Per-change verification is inspired by
process-supervision work on stepwise verification
\citep{lightman2024verify}, and the judge role draws on
\citet{zheng2023judge}. The judge prompt makes three explicit distinctions:
geometry-tagged candidates indicate nearby vector linework but do not establish
its structural meaning; a move is not a deletion at the original location; and
unit conversions match within 0.1\%. The last provision allows the judge to
rederive $72'\text{-}0'' = 21{,}945.6$~mm rather than reject a rounding
difference.
The judge uses the provider's medium reasoning setting to support the ordered
checks in the review prompt.

\begin{table}[t]
\centering
\caption{Agentic-layer configuration in paper millimeters and model-space meters.}
\label{tab:agenticcfg}
\small
\begin{tabular}{@{}llp{0.44\linewidth}@{}}
\toprule
Constant & Value & Role \\
\midrule
\multicolumn{3}{@{}l}{\textit{Overlay rendering}} \\
Composition width & 1600 px, $\le 8$ px/mm & Drawing legible without oversized payloads \\
Halo width, opacity & $2.8\times$, $\alpha = 0.55$ & Drawn ink stays black inside a colored halo \\
Content crop margin & 42 mm paper & Keeps grid bubbles and two dimension chains \\
\midrule
\multicolumn{3}{@{}l}{\textit{Model invocation}} \\
Provider model & \texttt{gemini-3.5-flash} \citep{google2026gemini35flash} & Pinned pretrained model used in the reported runs \\
Sampling & Provider defaults & No temperature or top-$p$ override \\
Thinking, improve / judge & medium / medium & Reasoning setting for ordered review checks \\
Image resolution & high & Pinned provider media-resolution setting \\
Output cap & 8192 / 8192 tokens & Improve / judge \\
Operations per call, merged & $\le 12$, $\le 24$ & Canonicalization precedes the merged cap \\
Attempts, judge rounds & $\le 2$, $\le 2$ & Retry only if no op survived \\
Timeout, retries & 75 s, 3 attempts & Linear backoff $1.5\,\mathrm{s} \times n$ \\
Accepted finish reason & \texttt{STOP} only & Truncated and incomplete responses fail closed \\
\midrule
\multicolumn{3}{@{}l}{\textit{Admission guards}} \\
Ink coverage, member & $\ge 0.55$ of 15 samples & Stroke within 2.4 mm paper of the axis \\
Glyph ink, column & 6 mm paper radius & Polygon centroid, circle $r \le 12$ mm, or $\ge 2$ strokes $\le 12$ mm \\
Endpoint snap & 0.75 m & To column centers and grid intersections \\
Duplicate endpoints & 0.35 m & 0.40 m for columns \\
Multi-bay coverage & reject at $\ge 0.5$ & Rival member within 0.3 m over 15 samples \\
Minimum member length & 0.4 m & After snapping \\
Section assignment & blank target, exact local text & Existing sections are immutable \\
\midrule
\multicolumn{3}{@{}l}{\textit{Destructive guards}} \\
Deletion cap per class & $\max(3,\ \lceil 0.4 N\rceil)$ & $N$ from the pre-pass count \\
Displaced-column refusal & 0.3--1.5 m & Inked free intersection near the mark \\
At-mark glyph radius & 3 mm paper & Tight: a correct mark is centered \\
Occupancy radius & 0.45 m & Candidate intersection must be free \\
Move limits & 0.05--3 m & Below is treated as no change; above is rejected \\
Contradiction & move then delete & Rejected within one pass \\
\midrule
\multicolumn{3}{@{}l}{\textit{Metric guards}} \\
Calibration gate & assumed or conf. $< 0.65$ & Never overrides high-confidence consensus \\
Factor clamp, dead band & $[0.2, 5]$, $\ge 2\%$ & Calibration is an exclusive transaction \\
Grid pair separation & $\ge 100$ mm & Short pairs amplify error \\
\midrule
\multicolumn{3}{@{}l}{\textit{Deterministic candidate generators}} \\
Unmarked column & free intersection, glyph ink & No column within 0.45 m, no wall within 0.4 m \\
Unmarked span & $\ge 0.6$ ink, span 1.2--20 m & Existing coverage $< 0.35$; typed by hatch \\
Unmarked joist & bay fraction repeated & $\ge 0.6$ ink, no joist within 0.3 m \\
Inkless beam/wall & $< 0.25$ axis coverage & Hatched walls exempt \\
Inkless column & no glyph within 6 mm paper & Candidate deletion; displaced-column guard still applies \\
Calibration seed & $\ge 3$ dimensions, $\ge 60\%$ within 2\% & Preferred over a model calibration proposal \\
\bottomrule
\end{tabular}
\end{table}

Failure containment begins with the visual prompt. Opaque centerlines can
obscure thin source strokes, whereas wide translucent halos preserve the
underlying linework for inspection. A failed improvement call does not terminate
the pass because the other call and deterministic seeds may still supply
candidates. The deterministic entity layout is returned unchanged when no
admitted entity operation survives, or when neither the complete batch nor a
reviewed subset is accepted. This fail-closed scope excludes printed floor-level
metadata, which can persist after only list-format and plausibility checks and
must therefore be confirmed by the user.

These guards constrain admissible changes but do not certify semantic
correctness. Structural additions require class-specific line or glyph evidence,
but deletions have no general absence-of-linework test, and a moved column need
not terminate at a detected glyph. Most attribute edits are validated through
identifiers, enumerated values, ranges, or exact nearby text rather than through
an independent semantic interpretation. For model-proposed calibration, neither
the physical dimension nor the free-text rationale is independently reread from
the drawing. A semantically incorrect operation can therefore pass both the
guards and the judge.

\subsubsection{Prompt architecture and reproducibility}
\label{sec:prompts}

The following boxes reproduce abridged implementation text with normalized
typography. Their capitalization and imperative wording are properties of the
runtime prompts rather than of the manuscript's narrative. Section-sign markers
identify omitted or substituted text. Because the omitted material can affect
model behavior, these excerpts document prompt structure but do not permit exact
replication. They correspond to protocol \texttt{gemini35-2026-08-13-v11} and
the pinned \texttt{gemini-3.5-flash} model. The excerpts include the
overlay legend prefixed to every call (Prompt~\ref{box:legend}), the improve
system prompt shared by both focused calls (Prompt~\ref{box:improve}), the two
focus suffixes that give those calls disjoint objectives
(Prompt~\ref{box:focus}), and the judge (Prompt~\ref{box:judge}). Three design
decisions structure the interaction. Checks are numbered and ordered to promote
consistent coverage; literal operation examples clarify required fields beyond
the provider schema; and each check states the expected negative outcome to
discourage unsupported corrections.

\begin{promptbox}[label={box:legend}]{Overlay legend, prefixed to every call}
\begin{lstlisting}[style=promptcompact]
Overlay legend: orange rectangles = columns, blue lines = girders, cyan = secondary beams, purple = shear walls, amber dashed = braces, green fill = slab regions, red fill = openings, gray dash-dot = grid lines with labels. The overlay is drawn as @@WIDE SEMI-TRANSPARENT HALOS@@: a marked member shows its dark drawn line inside a color halo; a drawn line with @@NO@@ color halo is @@UNMARKED@@ structure (a missed detection); a color halo over blank paper is a @@FALSE@@ detection.
\end{lstlisting}
\end{promptbox}

\begin{promptbox}[label={box:improve}]{The improve system prompt, shared by
both focused calls}
\begin{lstlisting}[style=promptcompact]
You are a senior structural engineer reviewing an automated markup of a structural/architectural floor-plan drawing. You receive ONE image (the original drawing with the current markup drawn on top) and the markup as JSON (coordinates in meters, y up). §§[overlay legend]§§

Work through these checks IN ORDER and propose at most 12 SMALL, HIGH-CONFIDENCE corrections as typed ops:

@@CHECK 0 - SCALE@@ (do this FIRST when the JSON says scale.method is "assumed" or scale.confidence < 0.65): The overlay always LOOKS aligned even when the scale is wrong - the error hides in the numbers. Compare the distance between two labeled grid lines in the JSON (their "position" values, meters) against what the drawing's printed dimension strings state for that same span. If they disagree by a consistent ratio, emit calibrate_scale with gridA, gridB and the stated real distance in mm. Never calibrate when the scale method is dimension-consensus with good confidence. When calibration is required, emit exactly ONE calibration op and NO other ops; review geometry and attributes in a fresh pass after calibration.

@@CHECK 1 - FALSE MARKS@@: an overlay element with NO dark drawing linework beneath it is a false detection - delete it. BRACES ARE SPECIAL: on a framing PLAN a vertical-bracing bay is marked only by a SMALL X or V symbol on the bay edge (often with an HSS label) - easy to miss at this zoom. If the JSON lists a brace and you see ANY small symbol or label anywhere along its span, keep it; deleting a brace should be EXTREMELY RARE and only when you are certain its span is completely bare.

@@CHECK 2 - MISSED STRUCTURE@@ (as valuable as check 1 - scan for it actively): every drawn beam centerline should carry a blue/cyan overlay, every drawn column glyph an orange rectangle, every thick/hatched wall band a purple band. Use the JSON "bayJoistCounts" table: neighboring bays normally repeat the same joist count, so a bay listed with FEWER joists than its neighbors usually misses EXACTLY the difference (typically ONE line). Look at that bay in the image, find the specific drawn joist line carrying no color halo, and add THAT line only. NEVER add more beams than the count difference, and never add a beam you cannot point at as a drawn-but-unhaloed line. For anything drawn but unmarked, emit add_beam / add_column / add_wall, ALWAYS reusing coordinates of listed entities; @@never estimate coordinates from pixels@@.

@@CHECK 3 - WRONG PLACE@@: a marked column sitting visibly OFF its drawn glyph / grid intersection -> move_column with the correct x, y. Attached members follow.

@@CHECK 4 - ATTRIBUTES@@: set_region_kind (slab vs stair/elevator opening), set_wall_thickness (clearly wrong thickness), set_beam_kind (girder vs secondary).

@@CHECK 5 - SECTION LABELS@@: compare printed section designations with the JSON. Emit set_section only when the target has no section, and copy the designation verbatim. Never overwrite a nonempty section.

§§[The general operation rules are omitted; literal examples appear in the operation-schema excerpt.]§§
\end{lstlisting}
\end{promptbox}

\begin{promptbox}[label={box:focus}]{The two focus suffixes, which give the
parallel calls disjoint objectives}
\begin{lstlisting}[style=promptcompact]
@@FOCUS OF THIS PASS (call A):@@ checks 1 and 3 ONLY (false marks to delete; misplaced columns to move). Do not add members or calibrate here. DISPLACED vs FALSE: an orange column mark floating near (within ~1 m of) a drawn column glyph or grid intersection that carries no other mark is a MISPLACED detection - move_column it onto the glyph; delete a column mark only when NO drawn glyph exists anywhere near it.

@@FOCUS OF THIS PASS (call B):@@ checks 0, 2, 4, 4b and 5 ONLY (scale calibration; drawn-but-unmarked structure; attributes; printed levels; section designations). Do not delete or move anything here. CHECK 0 (scale) is MANDATORY when scale is assumed or has low confidence. If calibration is required, return exactly one calibration op and defer all other checks to a fresh pass. For ADDITIONS: treat a bay whose bayJoistCounts value is strictly LOWER than its neighboring bays as a candidate for missing joists; add only a specific unhaloed structural line verified in the image. @@When the counts are uniform and no unhaloed linework exists, returning "ops": [] (or calibration only) is the normal outcome - never invent members.@@
§§[followed by the geometry-derived repair-candidate block]§§
\end{lstlisting}
\end{promptbox}

\begin{promptbox}[label={box:judge}]{The judge system prompt}
\begin{lstlisting}[style=promptcompact]
You are a strict reviewer of floor-plan markups. Image 1 shows the plan with markup A (original). Image 2 shows the SAME plan with markup B (revised). You also get the authoritative LIST of changes B applied - that list is exactly what changed; @@do NOT re-diff the two images for unlisted differences.@@ §§[overlay legend]§§

VERIFY EACH LISTED CHANGE against the drawing:

- "deleted <kind> <id>" is correct when Image 1 shows NO corresponding drawn structure under that mark - pale overlay color over blank paper. It is wrong when real linework or a bracing X/V symbol sits beneath. @@Deleting a column mark that merely sat NEAR its drawn glyph (a misplaced detection that should have been moved onto the glyph) is WRONG@@ - reject it via bad_changes.
- "added <kind>" is correct when the drawing shows that member drawn but unmarked: at the stated location, Image 1 shows a dark drawn line/glyph WITHOUT a color halo, and Image 2 shows the same line now carrying its halo. The authoritative list contains only changes already applied by the deterministic executor, so every listed addition is present in Image 2 by construction. Reject an addition only when its underlying mark is nonstructural or unsupported.
- An "ink scan" reason indicates that vector linework was detected near the candidate; it does not establish structural meaning. Reject the change when the evidence represents a dimension, annotation, or other nonstructural mark.
- "moved column <id> to (x, y)" means the SAME column left its old spot and now sits at the new one - @@never count its old location as a deletion.@@
- "calibrated scale" leaves the overlay unchanged - verify its grid distance against printed dimension strings; a conversion within @@0.1%@@ of the stated value matches.

Accept only when every listed change is supported. Otherwise reject and place exactly the wrong change numbers in @@bad_changes@@ so the remainder can be salvaged. Judge only the listed changes.
\end{lstlisting}
\end{promptbox}

\section{Benchmark and evaluation}
\label{sec:benchmark}

\begin{figure}[t]
\centering
\includegraphics[width=\linewidth]{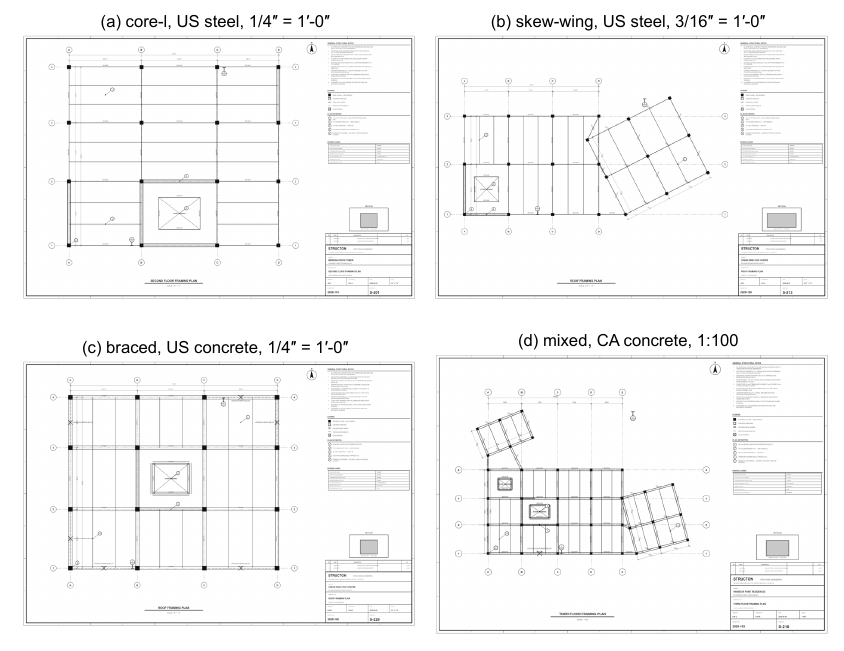}
\caption{Four \pdbench{} development drawings from the core-l, skew-wing,
braced, and mixed families.}
\label{fig:benchmark}
\end{figure}

The available public floor-plan datasets primarily annotate architectural
semantics \citep{kalervo2019cubicasa,fan2021floorplancad}, whereas prior
structural-drawing studies have evaluated small proprietary datasets
\citep{zhao2021reconstructing}. None of these datasets combines the structural
entity schema required here with exact model-space geometry. The resulting
benchmark therefore contains 100 procedurally generated single-story framing
plans in two equal halves with distinct roles. The \pdbench{} development half
supported all method development, and its failures informed every rule and
threshold revision. The \pdtest{} held-out half was generated from a disjoint
seed stream after the detection rules were frozen and was evaluated exactly
once. No rule, threshold, or prompt was changed in response to a held-out
result, and all reported recall and precision values derive from this half. Both
halves follow generator-defined United States and Canadian notation variants and include
full title blocks, revision strips, code-referenced general notes, legends,
keynote hexagons, design-load schedules, key plans, grid bubbles at both ends,
bay and overall dimension chains, section markers, and north arrows
(Figure~\ref{fig:benchmark}). Both halves were produced using the same
fixed-seed parametric drawing generator developed for the benchmark. Each family,
notation variant, and material position is matched case-for-case between the
halves, so the two differ only in their random parameter draws. The evaluation
release includes the drawings and ground truth but not the generator or
plan-to-model implementation. Automated checks found no inconsistencies in
either half for file completeness, count agreement, member degeneracy or
duplication, slab overlap, or coordinate mappings within page bounds. These
checks establish geometric and metadata consistency only; they do not establish
structural-design adequacy or compliance with every drafting standard.

Each half has the same distribution of notation variants across the three
materials. Of the 50 drawings in each half, 25 use the United States variant on ARCH-D
paper with imperial dimension strings and IBC, AISC,
ACI, and NDS note blocks, and 25 use the Canadian variant on
A1 paper with millimeter dimensions and NBCC, CSA S16, A23.3, and O86
notes. The materials comprise 17 steel, 17 concrete, and 16 timber cases.
Material selection determines the designation families, including
W14$\times$90 versus W360$\times$134, $18''\times 18''$ versus
$450\times450$, and glulam sizes in both notation variants. It also determines
single steel centerlines versus double lines at the true width for concrete and
glulam, concrete hatching versus cross-hatched timber walls, and
variant-consistent wall-thickness ranges. Each case includes a vector PDF, a
clean raster at 3.2~px/mm, a variant blurred with $\sigma = 1.5$, exact
$90^{\circ}$, $180^{\circ}$, and $270^{\circ}$ rotations of the blurred variant,
ground truth in model-space meters, and the complete world-to-sheet mapping.
The \pdtest{} ground truth totals 1{,}082 columns, 2{,}721 beams, 194 wall
panels, 47 braces, and 80 openings; the corresponding \pdbench{} totals are
1{,}100, 2{,}819, 189, 44, and 80.

Ten structural families target specific detector failure modes
(Table~\ref{tab:families}). Each family contributes five seeded variants to
\pdbench{} and five additional seeded variants to \pdtest{};
Appendix~\ref{app:inventory} lists the 50 held-out cases with their scales and
entity counts. The family definitions vary footprint shape, framing density,
skew, openings, walls, and bracing, which permits aggregate results to be
related to known geometric conditions. Because both halves are sampled from the
same parametric families, the held-out half measures generalization to unseen
drawings from the same distribution rather than to independent drafting
conventions. Neither half captures the full diversity of structural drawings or
design-office practices.

\begin{table}[t]
\centering
\caption{Ten \pdbench{} families and the capabilities they test.}
\label{tab:families}
\small
\begin{tabular}{@{}lp{0.72\linewidth}@{}}
\toprule
Family & Designed to test \\
\midrule
core-l & L-, T-, and C-shaped shear-wall cores and perimeter panels on regular grids: wall symbols, walls replacing beams, and variable-thickness wall geometry \\
core-u & Twin U-, box-, and C-cores with variable bay counts: multi-core layouts and core-adjacent framing \\
skew-wing & A wing rotated $14$--$28^{\circ}$ with sloped connector beams and rotated dimension chains: rotated lattices and skew-aware scale \\
chamfer & Chamfered corners producing diagonal edge beams and trimmed slabs \\
atrium & A multi-bay central void with corner diagonals: large-opening semantics and building-scale X symbols \\
braced & Up to six X-braced perimeter bays: brace signs against beam marks on shared edges \\
foot-l & L-shaped footprints with notches and stair and shaft openings \\
foot-u & U-shaped footprints with twin cores and edge bracing \\
dense & Bays of 4.5--10~m with two to four infill beams per bay: infill recall under density \\
mixed & Two skewed wings, a core, a stair, and bracing on one drawing: every class at once \\
\bottomrule
\end{tabular}
\end{table}

The generator represents dimensions, hatches, and opening boundaries using the
same primitive types processed by the extractor. Dimensions consist of rotated
text, extension lines, and tick strokes; hatches are explicit linework; and
openings remove interior framing and include perimeter trimmers. This controlled
representation exercises the intended input path. Although the held-out half
prevents tuning to particular drawings, the generator and extractor retain
shared drafting assumptions. Consequently, results within this corpus may be
optimistic for drawings from independent producers. As discussed in
Section~\ref{sec:conclusions}, evaluation on independently drafted drawings is
necessary to assess external generalization.

\subsection{Evaluation protocol}
\label{sec:evaluation}

All rule, threshold, prompt, and guard revisions were completed on \pdbench{}
before \pdtest{} was generated. The held-out half was then processed exactly
once per condition, and no observed held-out failure was used to modify the
system. Each held-out case passed once from its vector PDF through the complete
framework. Deterministic extraction and detection produced the intermediate
layout, and a single stochastic execution of guarded agentic refinement
produced the final layout scored by the translation-fitting and entity-scoring
procedures described below. The clean, blurred, and rotated raster variants are
provided for future studies of blur and rotation but are excluded from the
quantitative results. For scale, the relative error for each case is
$e_s=\lvert\hat{s}-s_{\mathrm{ref}}\rvert/s_{\mathrm{ref}}$, where
$s_{\mathrm{ref}}$ is read from the generator metadata. The estimated scale,
resolution method, confidence, and $e_s$ are retained with each evaluated case.
Before scoring, the procedure fits a translation using the median offset among
nearest column pairs; the results therefore do not evaluate recovery of the
absolute drawing origin. Columns, walls, braces, and openings use greedy
one-to-one matching after all eligible pairs have been sorted by increasing
geometric cost. Columns match within 0.5~m of center, walls
within 0.8~m at both endpoints, braces within 1.0~m at both endpoints, and
openings within 1.5~m of centroid. Wall thickness is not part of the reported
match predicate. These localization tolerances are permissive relative to member
dimensions, and the protocol does not test node connectivity, section or
material assignment, slabs, supports, releases, or solver validity.

Class-level recall and precision constitute the primary evaluation outcomes. A
secondary per-drawing regression gate requires column recall and
precision of at least 0.95, beam recall of at least 0.85, wall recall of at least
0.75, and opening recall of at least 0.50. This compound gate does not constrain
beam, wall, or opening precision and does not include braces, so its pass count
cannot be interpreted as a complete measure of accuracy. A rejected refinement
transaction is scored using the unchanged intermediate layout, so a case that
fails closed still produces an evaluation result. Internal acceptance indicates
only that a guarded transaction passed its checks; the external entity metrics
determine whether the revised layout is more accurate.

A separate controlled study applied two fixed corruptions to three development
drawings, with three repeated executions for each drawing. The member-repair
scenario inserted four
fabricated marks, removed predetermined structural entities, and displaced one
column. A strict pass required an accepted transaction, removal of all fabricated
marks, restoration of every removed entity, correction of the displaced column,
complete one-to-one recovery of the baseline entity collections, no collateral
or unexpected entities, and no changes to attributes of entities that persisted
from the corrupted input. Complete recovery required agreement within 0.20~m
for column centers and member endpoints, a Hausdorff distance no greater than
0.02~m for region boundaries, and agreement within 0.01~m for grid positions and
anchors. The calibration corruption
multiplied scale-sensitive layout coordinates, physical-width fields, curve
radii, region and grid geometry, and transform scale factors by $1/1.5$, then
marked the scale as assumed; parsed dimension records were left unchanged. Its
strict criterion required an accepted transaction, a primary grid-span error
below 0.1\%, restoration of every audited scale-sensitive field within the
implemented absolute and relative tolerances, complete entity recovery,
consistent derived
statistics, and no unrelated metadata or attribute change. Each scenario also
required complete and successful proposal and judge records that identified the
pinned model and reported a \texttt{STOP} finish reason. Because the same three drawings
and corruptions were reused, the 18 executions measure repeatability for those
cases rather than accuracy across 18 independent drawings.

\begin{figure}[t]
\centering
\includegraphics[width=0.72\linewidth]{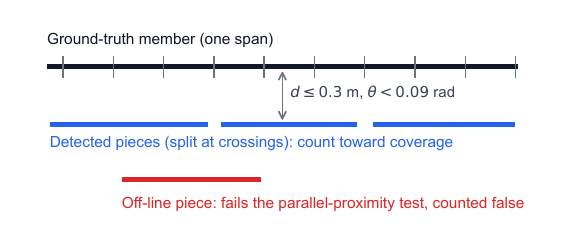}
\caption{Coverage-based beam scoring. Black denotes a ground-truth span, blue
denotes supporting normalized detections, and red denotes an off-axis false
detection.}
\label{fig:metric}
\end{figure}

Figure~\ref{fig:metric} illustrates the distinct coverage criteria used to score
beams. Before scoring, detected beam runs are normalized by splitting them at
detected columns and at interior crossings with other detected beams; pieces
shorter than 0.5~m are discarded. This normalized detection set is denoted by
$\mathcal{D}$ in Eq.~\eqref{eq:coverage}. The procedure accommodates a correct
drawn run that is represented as several per-bay pieces, or the converse,
without treating that representation difference as an error.

A ground-truth member $g$, sampled at $n = 10$ stations, is counted for
recall when
\begin{equation}
\mathrm{cov}(g) \;=\; \frac{1}{n} \sum_{k=1}^{n}
\mathbf{1}\!\left[\exists\, t \in \mathcal{D}: \operatorname{dist}(x_k, t)
\le 0.3~\text{m} \;\wedge\; \angle(g, t) < 0.09~\text{rad}\right] \;\ge\; 0.85,
\label{eq:coverage}
\end{equation}
where $\mathcal{D}$ is the normalized detection set described above. With ten
samples, the 0.85 threshold requires at least nine hits. A detected piece
contributes to the precision numerator when at least 0.70 of its own ten samples
lies
within 0.3~m and 0.09~rad of ground-truth members. Beam recall and precision
therefore have different numerators, which are reported separately in
Table~\ref{tab:aggregate}.

Recall is the fraction of ground-truth entities accounted for by the detector,
and precision is the fraction of detections supported by a ground-truth entity:
\begin{equation}
\mathrm{R} = \frac{\lvert \{\, g \in \mathcal{G} : g \text{ is matched} \,\} \rvert}{\lvert \mathcal{G} \rvert},
\qquad
\mathrm{P} = \frac{\lvert \{\, d \in \mathcal{D} : d \text{ is supported} \,\} \rvert}{\lvert \mathcal{D} \rvert},
\label{eq:rp}
\end{equation}
where $\mathcal{G}$ and $\mathcal{D}$ denote the ground-truth and detected
entities of a class, respectively. Beam recall and precision use the sampled
coverage tests above and therefore have separate hit counts. For columns,
walls, braces, and openings, matching is greedy and one-to-one within each
class; one detection cannot satisfy two ground-truth entities.

\section{Results}
\label{sec:results}

\subsection{End-to-end results on the held-out \pdtest{} half}
\label{sec:detresults}

The aggregate and disaggregated results follow the scoring definitions in
Section~\ref{sec:evaluation} and represent the complete framework: deterministic
extraction and detection followed by one stochastic execution of guarded
agentic refinement for each drawing. The held-out half was evaluated once. Beam
hit counts are reported separately for recall and precision because the coverage
metric permits a detected run to be split into several pieces, or multiple
per-bay pieces to be merged into a single run. Such representation differences
are not treated as errors when the sampled geometry satisfies the stated
thresholds.

\begin{figure}[t]
\centering
\includegraphics[width=0.66\linewidth]{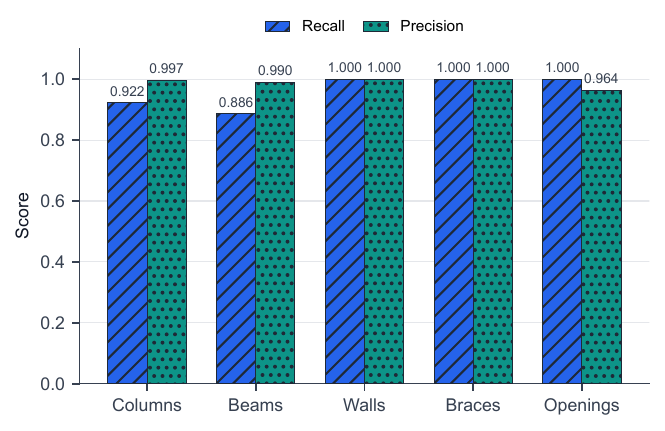}
\caption{Per-class recall and precision of the complete framework on the
held-out \pdtest{} half.}
\label{fig:results}
\end{figure}

Table~\ref{tab:aggregate} and Figure~\ref{fig:results} summarize the end-to-end
results. The scale-resolution stage produced dimension-consensus estimates with
a confidence of 1.0 for all 50 held-out drawings, and the maximum relative error
against the generator reference scale was 0.086\%. Recall and precision were
both 1.000 for the 194 wall panels and the 47 braces. Openings achieved a recall
of 1.000 and a precision of 0.964. Column recall was 0.922 with a precision of
0.997; three false columns remained on one drawing. Beam recall was 0.886 and
precision was 0.990, with 2,390 qualifying pieces among 2,413 normalized
detections. The wall metric reflects endpoint matching but not wall thickness,
whereas opening matching uses centroid proximity and does not assess boundary
agreement.

\begin{table}[t]
\centering
\caption{End-to-end entity recovery on the held-out \pdtest{} half, with
separate beam hit counts for recall and precision.}
\label{tab:aggregate}
\small
\begin{tabular}{@{}lrrrr@{}}
\toprule
Class & Recall & Precision & Recall hits / ground truth & Precision hits / detections \\
\midrule
Columns  & 0.922 & 0.997 & 998 / 1{,}082 & 998 / 1{,}001 \\
Beams    & 0.886 & 0.990 & 2{,}412 / 2{,}721 & 2{,}390 / 2{,}413 \\
Walls    & 1.000 & 1.000 & 194 / 194 & 194 / 194 \\
Braces   & 1.000 & 1.000 & 47 / 47 & 47 / 47 \\
Openings & 1.000 & 0.964 & 80 / 80 & 80 / 83 \\
\bottomrule
\end{tabular}
\end{table}

\begin{figure}[t]
\centering
\includegraphics[width=\linewidth]{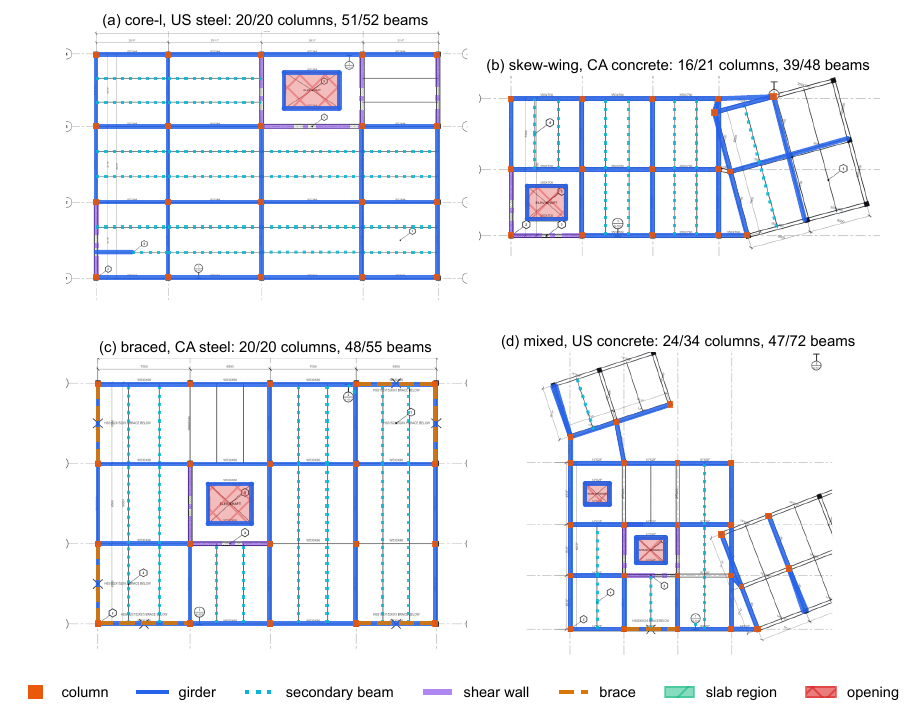}
\caption{Representative end-to-end overlays for four held-out \pdtest{} drawings.}
\label{fig:overlaymain}
\end{figure}

Figure~\ref{fig:overlaymain} presents representative correct recoveries and
recurring errors from the held-out half. Panels (b) and (d) show missed columns
and members in the rotated wings of the skew-wing and mixed families. These
overlays are consistent with the lower column and beam recall measured for the
rotated families, but they do not establish orientation as the cause. Of the 84
missed held-out columns, 80 occur in the mixed and skew-wing families, while the
remaining four occur in atrium drawings. Panel titles report the number of raw
detections divided by the ground-truth count, rather than the number of matched
detections divided by the ground-truth count.
Appendix~\ref{app:overlays} provides all 50 held-out overlays so that the
location and type of each recorded error can be inspected.

Table~\ref{tab:families_results} disaggregates the held-out results by family.
Seven of the ten families achieved a column recall of 1.000. The mixed (0.659)
and skew-wing (0.790) families contain wings rotated by 15 to $27^{\circ}$ and
account for nearly the entire column-recall deficit; the atrium family (0.969)
accounts for the remaining four missed columns. Beam recall was also lower for
the mixed (0.597), skew-wing
(0.858), and chamfer (0.873) families. The overlays locate missed detections in
rotated wings and at sloped connectors, but they do not identify a unique causal
mechanism. Column precision was 1.000 in every family except skew-wing (0.965).
The aggregate values also obscure a localized pattern in opening precision,
which decreased to 0.769 for the atrium family, where corresponding errors had
occurred in the development half. Visual inspection located these errors near
large X-shaped plan features, but the evaluation does not isolate the
contribution of individual mechanisms.

\begin{table}[t]
\centering
\caption{Per-family end-to-end recall and precision on the held-out half;
``--'' denotes an undefined metric because its denominator is zero.}
\label{tab:families_results}
\small
\begin{tabular}{@{}lcccccccccc@{}}
\toprule
& \multicolumn{2}{c}{Columns} & \multicolumn{2}{c}{Beams}
& \multicolumn{2}{c}{Walls} & \multicolumn{2}{c}{Braces}
& \multicolumn{2}{c}{Openings} \\
\cmidrule(lr){2-3}\cmidrule(lr){4-5}\cmidrule(lr){6-7}\cmidrule(lr){8-9}\cmidrule(l){10-11}
Family & R & P & R & P & R & P & R & P & R & P \\
\midrule
atrium & 0.969 & 1.000 & 0.939 & 0.993 & 1.000 & 1.000 & -- & -- & 1.000 & 0.769 \\
braced & 1.000 & 1.000 & 0.893 & 0.995 & 1.000 & 1.000 & 1.000 & 1.000 & 1.000 & 1.000 \\
chamfer & 1.000 & 1.000 & 0.873 & 1.000 & 1.000 & 1.000 & -- & -- & 1.000 & 1.000 \\
core-l & 1.000 & 1.000 & 0.982 & 1.000 & 1.000 & 1.000 & -- & -- & 1.000 & 1.000 \\
core-u & 1.000 & 1.000 & 0.971 & 1.000 & 1.000 & 1.000 & -- & -- & 1.000 & 1.000 \\
dense & 1.000 & 1.000 & 0.909 & 0.993 & 1.000 & 1.000 & 1.000 & 1.000 & 1.000 & 1.000 \\
foot-l & 1.000 & 1.000 & 0.984 & 1.000 & 1.000 & 1.000 & -- & -- & 1.000 & 1.000 \\
foot-u & 1.000 & 1.000 & 0.971 & 0.975 & 1.000 & 1.000 & 1.000 & 1.000 & 1.000 & 1.000 \\
mixed & 0.659 & 1.000 & 0.597 & 0.978 & 1.000 & 1.000 & 1.000 & 1.000 & 1.000 & 1.000 \\
skew-wing & 0.790 & 0.965 & 0.858 & 0.975 & 1.000 & 1.000 & -- & -- & 1.000 & 1.000 \\
\bottomrule

\end{tabular}
\end{table}

Two descriptive partitions report variation within the corpus by notation
variant and material (Table~\ref{tab:cuts}). Because notation labels alternate
among drawings emitted by the same generator, similar results indicate balance
within this synthetic corpus rather than invariant transfer across independently
drafted conventions. On the held-out half, column recall was 0.933 for the 25
United States drawings and 0.912 for the 25 Canadian drawings. By material,
column recall was 0.926 for steel, 0.917 for concrete, and 0.924 for timber.
Timber drawings retained lower opening precision (0.897), consistent with the
development-half pattern, while refinement increased their column precision to
0.991. These aggregate partitions cannot attribute individual errors to a
single predicate, and dense joist infill remains only a possible contributor to
the beam result.

\begin{table}[t]
\centering
\caption{Held-out end-to-end results by notation variant and material; $n$ is
the number of drawings and ``--'' denotes an undefined metric.}
\label{tab:cuts}
\small
\setlength{\tabcolsep}{4pt}
\begin{tabular}{@{}lccccccccccc@{}}
\toprule
& & \multicolumn{2}{c}{Columns} & \multicolumn{2}{c}{Beams}
& \multicolumn{2}{c}{Walls} & \multicolumn{2}{c}{Braces}
& \multicolumn{2}{c}{Openings} \\
\cmidrule(lr){3-4}\cmidrule(lr){5-6}\cmidrule(lr){7-8}\cmidrule(lr){9-10}\cmidrule(l){11-12}
Cut & $n$ & R & P & R & P & R & P & R & P & R & P \\
\midrule
US & 25 & 0.933 & 0.994 & 0.903 & 0.987 & 1.000 & 1.000 & 1.000 & 1.000 & 1.000 & 1.000 \\
CA & 25 & 0.912 & 1.000 & 0.869 & 0.994 & 1.000 & 1.000 & 1.000 & 1.000 & 1.000 & 0.930 \\
\midrule
steel & 17 & 0.926 & 1.000 & 0.883 & 0.986 & 1.000 & 1.000 & 1.000 & 1.000 & 1.000 & 1.000 \\
concrete & 17 & 0.917 & 1.000 & 0.880 & 0.995 & 1.000 & 1.000 & 1.000 & 1.000 & 1.000 & 1.000 \\
timber & 16 & 0.924 & 0.991 & 0.897 & 0.991 & 1.000 & 1.000 & 1.000 & 1.000 & 1.000 & 0.897 \\
\bottomrule

\end{tabular}
\end{table}

During held-out evaluation, the refinement layer accepted 43 of 50 guarded
transactions and applied 338 typed operations: 218 deletions, 70 member
additions, and 50 section-designation assignments. The remaining seven
transactions failed closed, leaving their intermediate layouts unchanged. The
designation assignments provide section and material semantics for the model
builder, although the entity-presence metrics do not assess those attributes.
The same operation channel also supports scale calibration when deterministic
resolution is unavailable. The function of the refinement layer therefore
extends beyond the entity counts in Table~\ref{tab:aggregate}.
Provider metadata verified use of the pinned \texttt{gemini-3.5-flash} model in
all 176 transport records, each of which returned a successful response. Only
responses with a \texttt{STOP} finish reason could contribute edits; ten
responses that reached the output-token limit failed closed. The secondary
per-drawing gate of Section~\ref{sec:evaluation} passed on 38 of the 50
drawings. Because this compound gate omits several class-precision criteria and
all brace criteria, the count provides only secondary evidence. An end-to-end
execution on the development half, retained in the data release, provides the
within-distribution reference. No held-out value was more than 2.4 percentage
points below its development counterpart, and beam recall accounted for the
largest difference. The framework therefore showed limited degradation on
unseen drawings from the same distribution. Because the refinement stage is
stochastic, each reported value reflects one execution per drawing and has no
repeatability interval. The run also evaluates the complete v11 protocol rather
than an isolated model choice, so it cannot separate the effects of the model,
prompts, image resolution, candidate generators, guards, and judging procedure.

\subsection{Controlled corruption repeatability}
\label{sec:controlledresults}

The controlled study produced 14 strict passes among 18 scenario executions
(Table~\ref{tab:controlled}). All nine calibration repetitions passed, the
largest postcalibration relative error in the primary span was 0.00182\%, and
all 2{,}073 checks of full-layout scale restoration passed. The member-repair
scenario satisfied every strict criterion in five of nine repetitions. Complete
one-to-one recovery of the baseline entity collections occurred in six of nine
repetitions. One of these six did not satisfy the complete strict criterion only
because eight existing slab regions were assigned a thickness of 120~mm where
the input value had been undefined.

\begin{table}[t]
\centering
\caption{Strict outcomes for three repeated controlled-corruption trials per drawing.}
\label{tab:controlled}
\small
\begin{tabular}{@{}lccc@{}}
\toprule
Drawing & Member repair & Complete entity recovery & Calibration \\
\midrule
\texttt{st-us-01} & 0/3 & 0/3 & 3/3 \\
\texttt{co-us-02} & 3/3 & 3/3 & 3/3 \\
\texttt{co-ca-03} & 2/3 & 3/3 & 3/3 \\
\midrule
Aggregate & 5/9 & 6/9 & 9/9 \\
\bottomrule
\end{tabular}
\end{table}

Across the repeated member-corruption executions, the refiner removed all 36
fabricated marks and corrected all nine displaced columns. It restored six of
nine deleted columns and all six deleted walls, but it restored none of the six
deleted beams. No collateral deletions or unexpected entities were observed.
Every scenario produced an internally accepted transaction, reinforcing the
distinction between transaction acceptance and strict end-state recovery.
Provider metadata verified the pinned model and a \texttt{STOP} finish reason
for every eligible controlled response. These results characterize repeated
behavior on three selected development drawings and do not estimate accuracy
for a broader drawing population.

\FloatBarrier
\section{Conclusions}
\label{sec:conclusions}

The proposed hybrid workflow converts structural framing-plan PDFs, supplied as
CAD exports or image-only files, into editable finite-element model drafts. To
the authors' knowledge, it is
the first workflow to apply an agentic vision-language layer directly to
building-component detection and structural model drafting from drawings. The
deterministic layer extracts primitives, resolves scale by dimension consensus,
identifies structural entities through an explicit drafting grammar, and
assembles their geometry using bearing topology. The guarded agentic layer
addresses residual ambiguities through typed operations, geometry-derived
candidates, operation-specific admission tests, strict change-level review, and
fail-closed transactions. This allocation of responsibilities preserves an
inspectable geometric core while restricting the actions available to the
pretrained vision-language model. The output remains a reviewable model draft,
not an analysis-ready structural model.

The two benchmark halves separate method development from performance
measurement. The \pdbench{} half informed every rule and threshold revision,
whereas the seed-disjoint \pdtest{} half was generated after the rules were
frozen and evaluated once. On the held-out half, the complete framework
estimated scale within 0.1\% of the generator reference for every drawing.
End-to-end recall and precision were 0.922 and 0.997 for columns and 0.886 and
0.990 for beams, respectively. Walls, braces, and openings achieved a recall of
1.000, with respective precision values of 1.000, 1.000, and 0.964. No held-out
value was more than 2.4 percentage points below its development-half
counterpart, indicating limited degradation for unseen drawings from the same
distribution. Missed columns and beams were concentrated in families containing
rotated wings, skewed connectors, and diagonal boundaries, although the
experiments do not isolate a causal mechanism. Because refinement is stochastic,
each reported value represents one execution per drawing and has no
repeatability interval.

The controlled corruptions further characterize both capability and remaining
failure modes. Calibration passed all nine repetitions, with a maximum
primary-span error of 0.00182\% and no failures among 2{,}073 full-layout
restoration checks. Member repair satisfied every strict criterion in five of
nine repetitions. All 36 fabricated marks were removed, all nine displaced
columns were corrected, and all six deleted walls were restored, but none of the
six deleted beams was recovered. Complete baseline-entity recovery within the
controlled tolerances occurred in six of nine repetitions; one of these
executions failed the overall strict criterion only because it assigned
thickness to eight slab regions whose previous value was undefined. All 18
transactions were accepted internally, confirming that guarded acceptance does
not imply correct end-state recovery. Provider metadata verified the reported
model and successful finish state, but the repeated observations involve only
three unique development drawings.

The term ``training-free'' denotes the absence of a task-specific detector
trained or fine-tuned on plan annotations. The complete system nevertheless
contains a learned component because refinement invokes a pretrained,
externally hosted vision-language model, and the deterministic rules reflect
manual design choices. The held-out half prevents tuning to particular drawings,
but the generator and detector share representational assumptions; held-out
performance may therefore be optimistic for drawings from independent
producers. Similar results for the encoded United States and Canadian variants
indicate balance within the generator, not transfer across design offices or
regional practices. Quantitative evaluation is limited to vector PDFs, while
the raster and model-assembly examples are illustrative. Furthermore, the fitted
translation and class-dependent localization tolerances do not assess wall
thickness, analytical connectivity, sections, materials, slabs, supports,
releases, load paths, solver validity, or agreement in structural response.
Published results for learned floor-plan and CAD parsers use different classes,
units of analysis, data splits, and matching rules, and thus provide context
rather than directly comparable baselines
\citep{kalervo2019cubicasa,fan2022cadtransformer,zhao2021reconstructing}.

Every generated draft requires engineering review before analysis. The
frozen-rule evaluation should be extended to independently drafted plans from
additional offices, regions, materials, renovation projects, and image-quality
conditions. Multiple executions per drawing are needed to estimate
repeatability, and ablation studies should separate the effects of deterministic
seeds, prompts, admission guards, and judging. Future evaluations should report
operation-level error rates alongside entity metrics and assess graph-level
connectivity, section and material fidelity, solver checks, and uncertainty.
A user study should also compare modeling time, correction effort, computation,
and provider-call cost with deterministic-only, learned, and manual workflows.
Within the stated limitations, deterministic geometry combined with guarded
vision-language review provides a transparent and testable path from structural
drawings to editable model drafts. Its value for independently drafted drawings
remains to be established.

\section*{Data availability}

The benchmark release contains all 100 vector drawings in the \pdbench{}
development half and the \pdtest{} held-out half. Each drawing is accompanied by
a clean raster, degraded variants, exact ground truth,
world-to-sheet-to-pixel mappings, and a proposed evaluation protocol that
preserves the development-test separation. The benchmark archive will be
deposited in a
public repository upon acceptance, and its persistent identifier will be added
to the published version. A sanitized evidence bundle accompanying the
manuscript reports per-drawing and controlled-trial outcomes,
whitelisted protocol settings, provider-audit counts, source hashes, and file
checksums. Complete successful provider responses and hashed request metadata
were retained in a restricted private audit archive. Prompts, raw provider
messages, images, layouts, and local paths are excluded from the shared bundle
for security and third-party service reasons. The release also excludes the
drawing generator and the source code for detection and model generation. The
available artifacts therefore support verification of the reported aggregates
but not exact recomputation of detections or end-to-end reproduction of the
workflow.

{\small
\bibliographystyle{unsrtnat}
\bibliography{references}
}

\clearpage
\appendix

\section{Case inventory and per-case recall counts}
\label{app:inventory}

Table~\ref{tab:inventory} lists every held-out \pdtest{} case by family,
notation variant, material, drawing scale, and per-class end-to-end recall count,
reported as the matched count over the ground-truth count. Held-out variant
numbers range from 06 to 10 within each family, while development variant
numbers range from 01 to 05. Their per-case records and the per-case records for
the deterministic arm are included in the data release. Scale abbreviations are
$1/4''$ for $1/4'' = 1'\text{-}0''$ (1:48), $3/16''$ for 1:64, and $1/8''$ for
1:96. The inventory presents the denominator and recall outcome for each drawing
in addition to the corpus-level aggregate.

{\footnotesize
\setlength{\tabcolsep}{4pt}
\renewcommand{\arraystretch}{0.92}
\begin{longtable}{@{}lllllccccc@{}}
\caption{Held-out \pdtest{} inventory and per-case end-to-end recall counts
(matched/ground truth).}
\label{tab:inventory}\\
\toprule
Case & Family & Notation & Material & Scale & Cols & Beams & Walls & Braces & Open. \\
\midrule
\endfirsthead
\multicolumn{10}{l}{\tablename~\thetable{} (continued)}\\
\toprule
Case & Family & Notation & Material & Scale & Cols & Beams & Walls & Braces & Open. \\
\midrule
\endhead
\endlastfoot
atrium-06 & atrium & US & timber & 3/16$''$ & 24/25 & 54/56 & 2/2 & 0/0 & 2/2 \\
atrium-07 & atrium & CA & steel & 1:75 & 24/25 & 53/56 & 2/2 & 0/0 & 2/2 \\
atrium-08 & atrium & US & concrete & 3/16$''$ & 24/25 & 53/56 & 2/2 & 0/0 & 2/2 \\
atrium-09 & atrium & CA & timber & 1:75 & 24/25 & 49/56 & 2/2 & 0/0 & 2/2 \\
atrium-10 & atrium & US & steel & 3/16$''$ & 30/30 & 66/69 & 2/2 & 0/0 & 2/2 \\
braced-06 & braced & CA & concrete & 1:75 & 20/20 & 47/53 & 4/4 & 6/6 & 1/1 \\
braced-07 & braced & US & timber & 1/4$''$ & 20/20 & 51/55 & 2/2 & 7/7 & 1/1 \\
braced-08 & braced & CA & steel & 1:75 & 20/20 & 50/55 & 2/2 & 6/6 & 1/1 \\
braced-09 & braced & US & concrete & 3/16$''$ & 20/20 & 48/54 & 3/3 & 6/6 & 1/1 \\
braced-10 & braced & CA & timber & 1:75 & 20/20 & 45/53 & 4/4 & 4/4 & 1/1 \\
chamfer-06 & chamfer & CA & steel & 1:75 & 17/17 & 38/44 & 4/4 & 0/0 & 1/1 \\
chamfer-07 & chamfer & US & concrete & 1/4$''$ & 17/17 & 42/44 & 4/4 & 0/0 & 1/1 \\
chamfer-08 & chamfer & CA & timber & 1:50 & 17/17 & 38/44 & 4/4 & 0/0 & 1/1 \\
chamfer-09 & chamfer & US & steel & 1/4$''$ & 17/17 & 36/44 & 4/4 & 0/0 & 1/1 \\
chamfer-10 & chamfer & CA & concrete & 1:75 & 17/17 & 38/44 & 4/4 & 0/0 & 1/1 \\
core-l-06 & core-l & US & steel & 3/16$''$ & 20/20 & 50/52 & 5/5 & 0/0 & 1/1 \\
core-l-07 & core-l & CA & concrete & 1:75 & 20/20 & 52/52 & 5/5 & 0/0 & 1/1 \\
core-l-08 & core-l & US & timber & 1/4$''$ & 16/16 & 39/40 & 4/4 & 0/0 & 1/1 \\
core-l-09 & core-l & CA & steel & 1:50 & 16/16 & 40/40 & 4/4 & 0/0 & 1/1 \\
core-l-10 & core-l & US & concrete & 1/4$''$ & 16/16 & 39/40 & 4/4 & 0/0 & 1/1 \\
core-u-06 & core-u & CA & timber & 1:75 & 20/20 & 40/43 & 7/7 & 0/0 & 2/2 \\
core-u-07 & core-u & US & steel & 3/16$''$ & 20/20 & 50/52 & 8/8 & 0/0 & 2/2 \\
core-u-08 & core-u & CA & concrete & 1:75 & 20/20 & 42/43 & 7/7 & 0/0 & 2/2 \\
core-u-09 & core-u & US & timber & 1/4$''$ & 20/20 & 52/52 & 8/8 & 0/0 & 2/2 \\
core-u-10 & core-u & CA & steel & 1:75 & 24/24 & 52/53 & 7/7 & 0/0 & 2/2 \\
dense-06 & dense & US & concrete & 3/16$''$ & 20/20 & 71/79 & 4/4 & 2/2 & 2/2 \\
dense-07 & dense & CA & timber & 1:75 & 20/20 & 72/79 & 4/4 & 2/2 & 2/2 \\
dense-08 & dense & US & steel & 3/16$''$ & 20/20 & 72/79 & 4/4 & 2/2 & 2/2 \\
dense-09 & dense & CA & concrete & 1:75 & 20/20 & 53/57 & 4/4 & 1/1 & 2/2 \\
dense-10 & dense & US & timber & 3/16$''$ & 20/20 & 51/57 & 4/4 & 1/1 & 2/2 \\
foot-l-06 & foot-l & US & steel & 3/16$''$ & 18/18 & 49/49 & 2/2 & 0/0 & 2/2 \\
foot-l-07 & foot-l & CA & concrete & 1:50 & 18/18 & 49/49 & 2/2 & 0/0 & 2/2 \\
foot-l-08 & foot-l & US & timber & 1/4$''$ & 18/18 & 49/49 & 2/2 & 0/0 & 2/2 \\
foot-l-09 & foot-l & CA & steel & 1:75 & 18/18 & 45/49 & 2/2 & 0/0 & 2/2 \\
foot-l-10 & foot-l & US & concrete & 1/4$''$ & 18/18 & 49/49 & 2/2 & 0/0 & 2/2 \\
foot-u-06 & foot-u & CA & timber & 1:75 & 22/22 & 55/55 & 6/6 & 1/1 & 2/2 \\
foot-u-07 & foot-u & US & steel & 3/16$''$ & 22/22 & 54/55 & 6/6 & 1/1 & 2/2 \\
foot-u-08 & foot-u & CA & concrete & 1:75 & 22/22 & 52/55 & 6/6 & 1/1 & 2/2 \\
foot-u-09 & foot-u & US & timber & 3/16$''$ & 22/22 & 54/55 & 6/6 & 1/1 & 2/2 \\
foot-u-10 & foot-u & CA & steel & 1:75 & 22/22 & 52/55 & 6/6 & 1/1 & 2/2 \\
mixed-06 & mixed & CA & steel & 1:100 & 22/34 & 41/72 & 4/4 & 1/1 & 2/2 \\
mixed-07 & mixed & US & concrete & 1/8$''$ & 24/34 & 44/72 & 4/4 & 1/1 & 2/2 \\
mixed-08 & mixed & CA & timber & 1:100 & 19/34 & 42/72 & 4/4 & 1/1 & 2/2 \\
mixed-09 & mixed & US & steel & 1/8$''$ & 23/34 & 44/72 & 4/4 & 1/1 & 2/2 \\
mixed-10 & mixed & CA & concrete & 1:100 & 24/34 & 44/72 & 4/4 & 1/1 & 2/2 \\
skew-wing-06 & skew-wing & US & concrete & 1/8$''$ & 17/21 & 42/48 & 2/2 & 0/0 & 1/1 \\
skew-wing-07 & skew-wing & CA & timber & 1:75 & 17/21 & 42/48 & 2/2 & 0/0 & 1/1 \\
skew-wing-08 & skew-wing & US & steel & 3/16$''$ & 17/21 & 42/48 & 2/2 & 0/0 & 1/1 \\
skew-wing-09 & skew-wing & CA & concrete & 1:100 & 16/21 & 40/48 & 2/2 & 0/0 & 1/1 \\
skew-wing-10 & skew-wing & US & timber & 3/16$''$ & 16/21 & 40/48 & 2/2 & 0/0 & 1/1 \\
\bottomrule

\end{longtable}
}

\section{Detection overlays for the held-out half}
\label{app:overlays}

\begin{figure}[p]
\centering
\includegraphics[width=\linewidth,height=0.94\textheight,keepaspectratio]{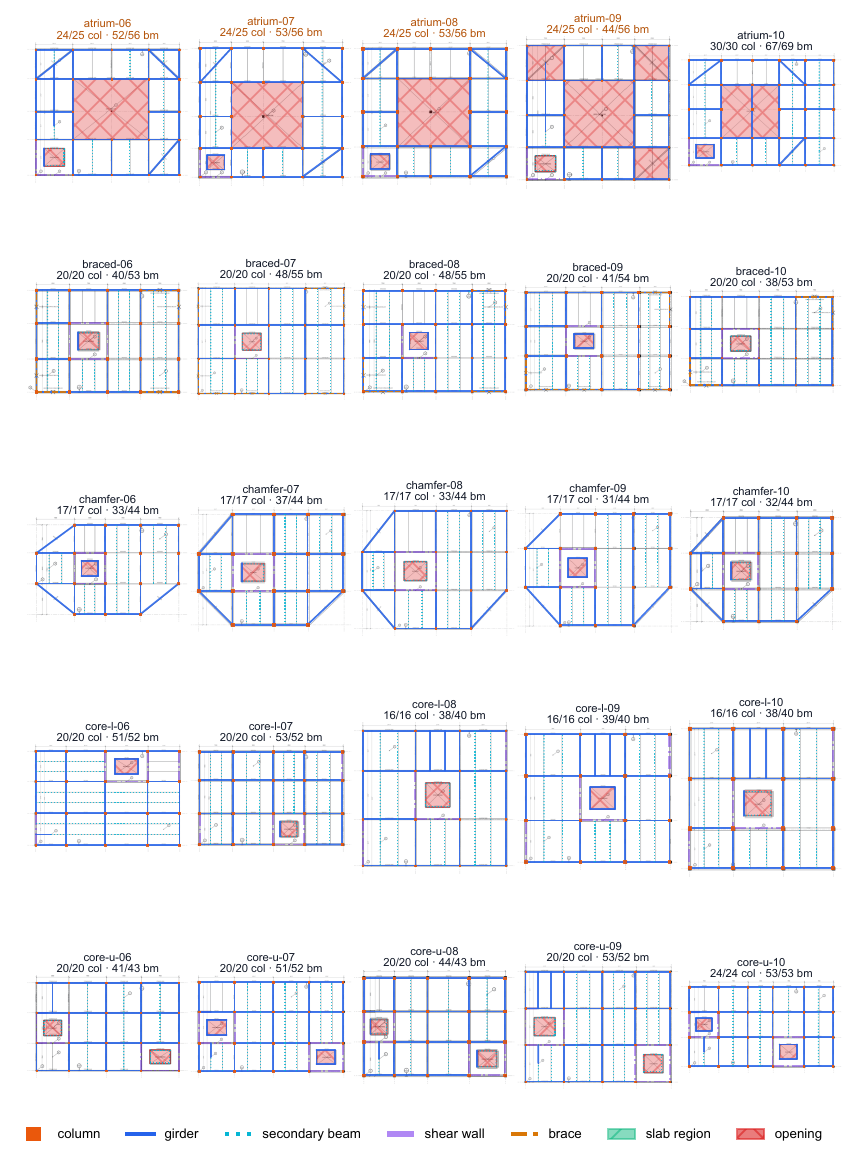}
\caption{Detection overlays, held-out cases 1--25 of \pdtest{}.}
\label{fig:overlayA}
\end{figure}

\begin{figure}[p]
\centering
\includegraphics[width=\linewidth,height=0.94\textheight,keepaspectratio]{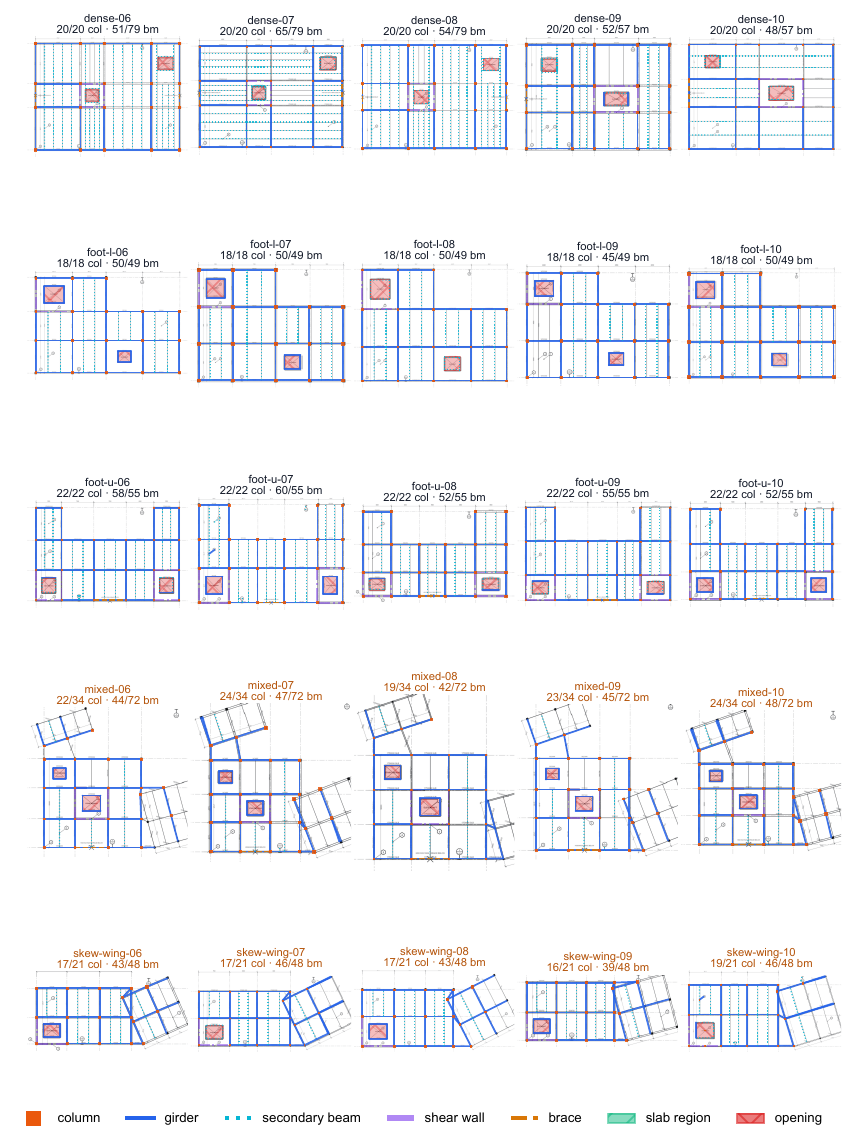}
\caption{Detection overlays, held-out cases 26--50 of \pdtest{}.}
\label{fig:overlayB}
\end{figure}

Figures~\ref{fig:overlayA} and~\ref{fig:overlayB} superimpose the final
end-to-end layout for every held-out \pdtest{} case on its source drawing, with
the view cropped to the structure. Counts below each panel compare the raw
detection count with the ground-truth count rather than reporting matched
detections. Equal counts can therefore conceal a false detection that offsets a
miss; the matched counts are reported in
Table~\ref{tab:inventory}. Titles are amber where the detected column count
differs from the ground-truth count. The errors discussed in
Section~\ref{sec:detresults} are visible as unmarked columns in the rotated wings
of the skew-wing and mixed families. These overlays permit case-level inspection
of the errors summarized in Tables~\ref{tab:aggregate}
and~\ref{tab:families_results}.

\end{document}